\documentclass[journal]{IEEEtran}

\usepackage{cite}
\usepackage[pdftex]{graphicx}
\usepackage{psfrag}			
\usepackage{amsmath}		
\usepackage{cancel}
\usepackage{amsfonts}		
\usepackage{url}				
\usepackage{here}				
\usepackage{color}			
\usepackage{balance}    
\usepackage{float}
\usepackage{bm}
\usepackage{bbm}
\usepackage{amsmath,amsfonts,amssymb}
\usepackage[]{algorithm2e}
\usepackage[utf8]{inputenc}
\usepackage{verbatim}
\usepackage{scalerel}
\usepackage{amssymb}
\usepackage{subfiles}
\usepackage{leftidx}
\usepackage[flushleft]{threeparttable}
\usepackage{subcaption}
\usepackage{hyperref}
\usepackage{enumitem}
\usepackage[font={small}]{caption}
\DeclareMathOperator*{\argmin}{\arg\!\min}

\begin{document}

\title{Representation-Free Model Predictive Control for Dynamic Motions in Quadrupeds\footnotemark}

\author{Yanran Ding$^1$,~\IEEEmembership{Student Member,~IEEE,}
        Abhishek~Pandala$^2$,~\IEEEmembership{Student Member,~IEEE,}
        Chuanzheng~Li$^1$,~\IEEEmembership{Student Member,~IEEE,}
        Young-Ha~Shin$^3$,~\IEEEmembership{Student Member,~IEEE,}
        and~Hae-Won~Park$^3$,~\IEEEmembership{Member,~IEEE}
\thanks{Manuscript received November 11, 2019; re-submitted July 16, 2020; revised November 23, 2020; accepted December 14, 2020. This paper was recommended for publication by Editor Eiichi Yoshida upon evaluation of the reviewers' comments. This work is supported in part by NAVER LABS Corp. under grant 087387, Air Force Office of Scientific Research under grant FA2386-17-1-4665, National Science Foundation under grant 1752262, the Mechanical Engineering Department of Korea Advanced Institute of Science and Technology (KAIST). \textit{(Corresponding author: Hae-Won Park)}, email: haewonpark@kaist.ac.kr}
\thanks{The preliminary version of this paper \cite{ding2019real} was presented in ICRA 2019.}
\thanks{Yanran Ding and Chuanzheng Li are with $^1$the Department of Mechanical Science and Engineering, University of Illinois at Urbana-Champaign, Urbana,IL, 61820 USA (e-mail: \{yding35, cli67\}@illinois.edu)}
\thanks{Abhishek Pandala is with $^2$the Department of Mechanical Engineering, Virginia Polytechnic Institute and State University, VA, 24061, USA (e-mail: agp19@vt.edu).}
\thanks{Young-Ha Shin and Hae-Won Park are with $^3$the Department of Mechanical Engineering, Korea Advanced Institute of Science and Technology, Daejeon-34141, South Korea (e-mail: \{shsin000,haewonpark\}@kaist.ac.kr)}
\thanks{This paper has supplementary downloadable multimedia material available at http://ieeexplore.ieee.org, provided by the authors. This material includes a video that presents the simulation and experiment results of the proposed MPC controller on a quadruped robot. Color versions of one or more of the figures in this paper are available online at http://ieeexplore.ieee.org.}
\thanks{Digital Object Identifier:}
\thanks{\textsuperscript{\textcopyright}2020 IEEE. Personal use of this material is permitted.  Permission from IEEE must be obtained for all other uses, in any current or future media, including reprinting/republishing this material for advertising or promotional purposes, creating new collective works, for resale or redistribution to servers or lists, or reuse of any copyrighted component of this work in other works.}
}

\markboth{IEEE TRANSACTIONS ON ROBOTICS}%
{Shell \MakeLowercase{\textit{et al.}}: Bare Demo of IEEEtran.cls for IEEE Journals}

\maketitle

\begin{abstract}
This paper presents a novel Representation-Free Model Predictive Control (RF-MPC) framework for controlling various dynamic motions of a quadrupedal robot in three dimensional (3D) space. Our formulation directly represents the rotational dynamics using the rotation matrix, which liberates us from the issues associated with the use of Euler angles and quaternion as the orientation representations. With a variation-based linearization scheme and a carefully constructed cost function, the MPC control law is transcribed to the standard Quadratic Program (QP) form. The MPC controller can operate at real-time rates of 250 Hz on a quadruped robot. Experimental results including periodic quadrupedal gaits and a controlled backflip validate that our control strategy could stabilize dynamic motions that involve singularity in 3D maneuvers.
\label{abstract}
\end{abstract}

\begin{IEEEkeywords}
Model Predictive Control, Legged Robots, Dynamics, Under-actuated Robots
\end{IEEEkeywords}

\IEEEpeerreviewmaketitle

\section{Introduction}
\label{sec:introduction}
\IEEEPARstart{T}{he} quadrupedal animals possess extraordinary competence of navigating harsh terrains by executing agile yet well-coordinated movements. For example, mountain goats demonstrate their extraordinary mobility on traversing steep cliffs \cite{aronson1982ibex}. Domesticated canine animals could be trained to execute a variety of acrobatic Parkour maneuvers \cite{dogParkour}. These remarkable abilities of quadrupedal animals motivated the development of many quadrupedal robots. Minitaur \cite{de2018vertical} realized various dynamic running gaits; ANYmal \cite{hutter2016anymal} and HyQ \cite{semini2011design} could navigate challenging terrains autonomously; MIT Cheetah robots achieved galloping \cite{seok2013design}, high speed bounding \cite{park2017high} and dynamic yet robust locomotion \cite{bledt2018cheetah}. As the capabilities of quadrupedal robot rapidly grow, related researches have geared towards motions beyond locomotion on flat terrains. For example, ANYmal demonstrated the stair climbing capability \cite{8460731}; MIT Cheetah 2 overcame obstacles by planning jumping trajectories online \cite{park2020jumping}; MIT Cheetah 3 achieved leaping onto high platforms \cite{8794449}; MIT Mini Cheetah could execute 360$^{\circ}$ backflips \cite{katz2019mini}. In general, the ability of quadrupedal robots is being developed towards applications that involve more dynamic maneuvers in increasingly complex scenarios.

\begin{figure}
	\centering
	\resizebox{0.8\linewidth}{!}{\includegraphics{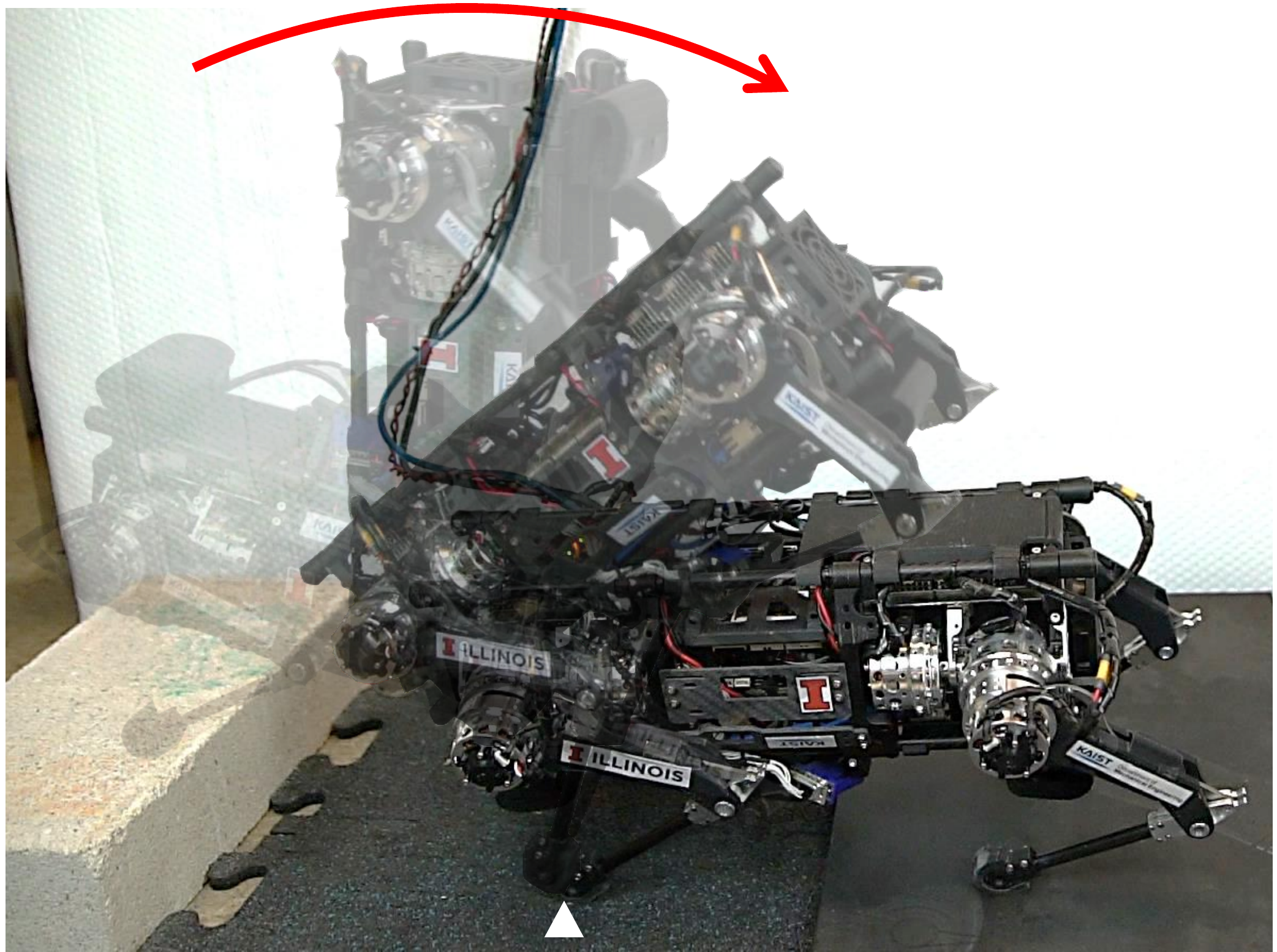}}
	\caption{Quadruped robot Panther performing a controlled backflip that involves passing through the upright pose, which corresponds to the singularity in the Euler angle representation. Throughout the controlled backflip, the feet pair indicated by the white triangle is kept in contact with the ground. The red arrow indicates the direction of the backflip and the shadowed images are snapshots during the backflip.}
	\label{fig:backflip}
\end{figure}

Controlling a robot to achieve similar mobility that rivals their animal counterparts encompasses many challenges. For instance, the control design should be capable of embracing the inherent dynamics when robots are under-actuated. In addition, the controller must take into consideration the constraints imposed by hardware capacity and the environment. Available solutions for dynamic locomotion include heuristic controller \cite{raibert1986legged}, inverse dynamics control\cite{righetti2013optimal} and hierarchical operational space control\cite{hutter2013hybrid}. Simulation result of backflip in Cassie is presented in \cite{xiong2020sequential}. Recent hardware results on ANYmal\cite{Hwangboeaau5872} demonstrated the potential of reinforcement learning in dynamic legged locomotion. Recent times have seen a surge in the use of optimization-based approaches, especially Model Predictive Control (MPC) for legged robots. Successful applications of MPC on humanoid\cite{herdt2010online}\cite{koenemann2015whole} and quadrupeds \cite{neunert2018whole} \cite{grandia2019frequency} have shown the efficacy of MPC in planning and controlling a wide variety of dynamic motions. 

Most of the MPC control frameworks in the quadrupedal robot community use Euler angles as the orientation representation \cite{feedbackMPC19, mastalli2020motion, passiveWBC20, Focchi2020} since many of locomotion tasks do not involve large deviation from the nominal orientation. However, Euler angles representation has the singularity issue \cite{shuster1993survey} (also known as Gimbal lock), which requires the motions to avoid singular orientations of the Euler angle representation. This disadvantage of Euler angles representation restricts the quadrupedal robot from executing motions such as climbing up near-vertical cliff as the mountain goat, or the acrobatic Parkour as the trained dogs. That is because these motions are highly likely to involve passing through the vicinity of singularity. Quaternion \cite{siciliano2016springer} is a singularity-free representation, but as mentioned in \cite{bhat1998topological}, quaternions have two local charts that covers the special orthogonal group $SO(3)$ twice. This ambiguity could cause unwinding phenomenon\cite{bhat1998topological}, where the body may start arbitrarily close to the desired attitude and yet rotate through large angles before reaching the desired orientation. Widely adopted by the Unmanned Aerial Vehicle (UAV) community, quaternions are often used in reactive controllers which instantaneously respond to the state of the vehicle. Sign function has been used in the reactive controller \cite{yang2019bee} to eliminate the ambiguity of the quaternion representation. In \cite{5991127}, hybrid dynamic algorithm has been introduced to solve the ambiguity of the quaternion representation. However, for predictive controllers such as MPC, switching of local charts is undesirable. The orientation of a rigid body is originally parameterized by the rotation matrix, which evolves on $SO(3)$ \cite{bullo2004geometric}. Although other orientation representations could be re-aligned to avoid their corresponding issues in a specific motion, the rotation matrix possesses advantages as a global parametrization that is compact and singularity-free.

In this manuscript we present the development and experiment results of a representation-free model predictive control (RF-MPC) framework that can be used to stabilize the robot in arbitrary orientation while leveraging the predictive capabilities of MPC.

\subsection{Contribution}
In this work, we venture to address the aforementioned problems in the following ways:
\begin{enumerate}[label=\arabic*)]
	\item We introduce a novel MPC formulation for controlling legged robots in dynamic 3D motions using rotation matrices. Specifically, a variation-based linearization technique\cite{wu2015variation}\cite{lee2011stable} is used to generate linear dynamics of the single rigid body (SRB) model, which leads to a representation-free MPC (RF-MPC) formulation with consistent performance even when executing motions that involve complex 3D rotations such as acrobatic motions in gymnastics that go through 90$^{\circ}$ pitch angle.
	\item The original orientation error involves the matrix logarithm map, which is a nonlinear function of the optimization variables. We propose an orientation error term in affine form of the state variables as an approximation to the original orientation error. This approximate orientation error term enables the MPC to be transcribed into a Quadratic Program (QP), which in term facilitates real-time MPC control.
	\item We implement the proposed RF-MPC on a quadruped robot Panther to realize real-time control of various gaits and a controlled backflip that passes through the singularity.
\end{enumerate} 

We have substantially built on the results of our previous work \cite{ding2019real} with the following novel strategies. (1) An improved formulation of the angular dynamics of the rotation matrix, which uses 3-dimensional variation vector instead of the 9-dimensional variation matrix. (2) A better choice of the objective function on the orientation which guarantees positive-definiteness. Previously, we used a special configuration error function \cite{bullo2004geometric} to approximate the orientation cost function, which only guarantees positive-definiteness when the predicted rotation matrix is on the $SO(3)$ manifold. (3) A section that provides a simulation case study on the effect of linearization schemes. (4) More experimental results, including trot running and the controlled backflip.

\subsection{Outline}
The paper is organized as follows: Section \ref{sec:MPC_formulation} presents the mathematical derivation of the RF-MPC framework. This section first introduces the single rigid body model, which is used for the derivation of the variation-based linearization of the dynamics. A novel vectorization technique along with a deliberately chosen cost function enables the transcription of RF-MPC into a Quadratic Program (QP). Section \ref{sec:Simulation} presents the numerical results, which includes various dynamic motions and a simulation case study that justifies the choice of the linearization scheme; Section \ref{sec:implement} summarizes the implementation details necessary for the application on the robot hardware platform; Section \ref{sec:Experiment} demonstrates the experimental results, followed by a discussion in Section \ref{sec:discussion}. Section \ref{sec:conclusion} provides the concluding remark with an outlook for the future work.

\section{Model Predictive Control}
\label{sec:MPC_formulation}

Model Predictive Control (MPC), also known as Receding Horizon Control (RHC), considers a model of the system to be controlled and repeatedly solves for the optimal control input subject to the state and control constraints. At each sampling time, a finite horizon optimal control problem is solved and the control signal for the first time-step is applied to the system during the following sampling interval. After that, the same process is repeated with the updated measurements. MPC-based controllers have the capability to incorporate various constraints that are essential to legged locomotion, including unilateral ground reaction force (GRF) and friction cone constraints. Besides, MPC could provide control laws that are discontinuous \cite{meadows1995receding}, which could not be easily achieved by conventional control techniques.

The MPC control law could be obtained by solving the following constrained optimization problem
\begin{subequations}\label{eq:nonlin_MPC}
	\begin{align}
		\text{minimize}& ~~~~~~ \ell_T(\bm{x}_{t+N|t}) + \sum_{k=0}^{N-1}\ell(\bm{x}_{t+k|t},\bm{u}_{t+k|t})\\
		\text{subject to}& ~~~~~~ \bm{x}_{t+k+1|t} = \bm{f}(\bm{x}_{t+k|t}) + \bm{g}(\bm{x}_{t+k|t})\bm{u}_{t+k|t}\\
		& ~~~~~~~~~~~~~~~ k=0,\cdots,N-1\\
		& ~~~~~~ \bm{x}_{t+k|t} \in \mathbb{X}, k=0,\cdots,N-1\\
		& ~~~~~~ \bm{u}_{t+k|t} \in \mathbb{U}, k=0,\cdots,N-1\\
		& ~~~~~~ \bm{x}_{t|t} = \bm{x}(t)=\bm{x}_{op}\\
		& ~~~~~~ \bm{x}_{t+N|t} \in \mathbb{X}_f
	\end{align}
\end{subequations}
where $\bm{x}\in\mathbb{R}^n,\bm{u}\in\mathbb{R}^m$ are the state and input vectors, respectively; $\ell_T:\mathbb{R}^n\rightarrow\mathbb{R}$ is the terminal cost function; $\ell:\mathbb{R}^{n}\times\mathbb{R}^{m}\rightarrow \mathbb{R}$ is the stage cost function; $N$ is the prediction horizon; $\bm{f}(\bm{x})+\bm{g}(\bm{x})\bm{u}$ is the control affine dynamic update equation; $\mathbb{X}\subseteq\mathbb{R}^n,\mathbb{U}\subseteq\mathbb{R}^m$ are the feasible polyhedral sets for the state and control; $\mathbb{X}_f$ is the final state set; $\bm{x}_{t+k|t}$ denotes the state vector at time $t+k$ predicted at time $t$, using the current state measurement $\bm{x}_{t|t}=\bm{x}_{op}$, where the subscript $(\cdot)_{op}$ denotes the variables at the current operating point. The operating point in this manuscript is defined as the current state $\bm{x}_{op}$ and control $\bm{u}_{op}$. 

In the case that the dynamic update equation $\bm{f}(\bm{x})+\bm{g}(\bm{x})\bm{u}$ is a nonlinear function, a nonlinear MPC (NMPC) could be formulated and solved as a general nonlinear program (NLP) by utilizing trajectory optimization (TO) techniques such as multiple shooting \cite{bock1984multiple} or direct collocation \cite{vonStryk1993}. 

Our main objective is to formulate a real-time executable MPC scheme for controlling quadruped robotics to perform a variety of dynamic motions. To meet the real-time requirement, the optimization problem posed by the MPC has to be solved robustly at a high rate on the mobile embedded computer, which has limited computational resources. Hence, a simplified model is adopted to reduce the dimensionality of the optimization problem. Since the mass of all legs combined is less than 10\% of the total body mass, a single rigid body model serves as a reasonable approximation.

\subsection{3D Single Rigid Body Model}
\label{sec:nonlin_dyn}	
\begin{figure}
	\centering
	\resizebox{0.85\linewidth}{!}{\includegraphics{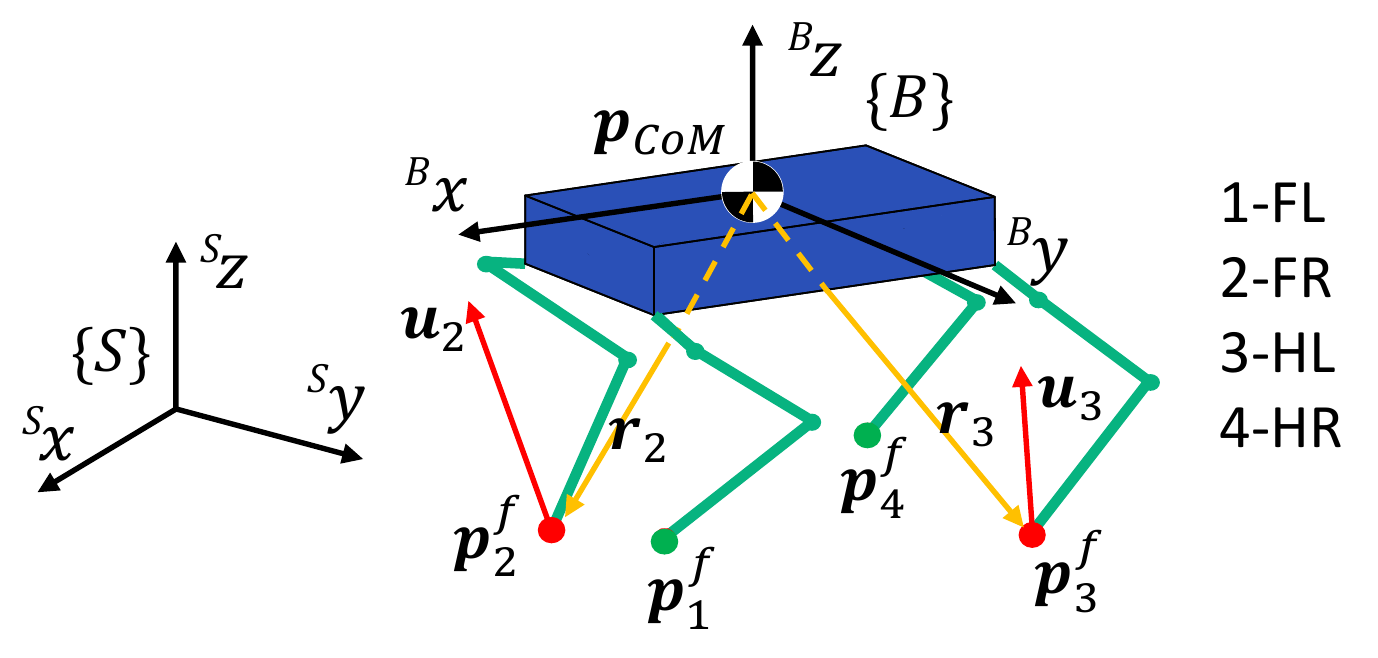}}
	\caption{Illustration of coordinate systems and the 3D single rigid-body model. $\{\bm{S}\}$ is the inertia frame and $\{\bm{B}\}$ is the body attached frame. $\bm{r}_i$ is the position vector from center of mass to each foot in $\{\bm{S}\}$ and $\bm{u}_i$ is the GRF of $i^{th}$ contact foot expressed in $\{\bm{S}\}$. The convention for the numbering of feet is such that FL stands for front-left leg, and HR stands for hind-right leg.}
	\label{fig:EoM}
\end{figure}

To mitigate the issue of demanding computational requirement of MPC for high Degrees of Freedom (DoF) system models, simple models or templates \cite{full1999templates} that capture the dominant system dynamics are used instead. Templates such as the Linear Inverted Pendulum \cite{kajita1991study} (LIP) is widely used in humanoid robots\cite{ramos2018humanoid} \cite{xiong2019orbit}. Centroidal dynamics \cite{Orin2013a} model is used in \cite{dai2014whole} \cite{wensing2013generation} \cite{li2020centroidal} to capture the major dynamic effect of the complex full-body dynamics model. The quadrupedal robot community has seen an increasing number of work that utilizes the SRB model in three-dimensional (3D) space, which assumes that the entire mass of the robot is lumped into a single rigid body (SRB). The simplicity of the SRB model is enabled by the light leg design, whose inertial effect is negligible compared with the body. Let the state of the single rigid body model be
\begin{equation}\label{state}
	\bm{x} := [\bm{p}~~\dot{\bm{p}}~~\bm{R}~~^B\bm{\omega}]\in\mathbb{R}^{18},
\end{equation}
where $\bm{p}\in \mathbb{R}^3$ is the position of the body Center of Mass (CoM); $\dot{\bm{p}}\in \mathbb{R}^3$ is the CoM velocity; $\bm{R}\in SO(3)=\{\bm {R}\in \mathbb{R}^{3\times 3}| \bm{R}^T \bm{R} = \mathbb{I}, {\rm det} (\bm{R}) = +1\}$ is the rotation matrix of the body frame $\{\bm{B}\}$ expressed in the inertial frame $\{\bm{S}\}$; $\rm det(\cdot)$ calculates the determinant of a matrix and $\mathbb{I}$ is the 3-by-3 identity matrix. Here, the rotation matrix $\bm{R}$ is reshaped into vector form. $^B\bm{\omega}\in \mathbb{R}^3$ indicates the angular velocity vector expressed in the body frame $\{\bm{B}\}$. Variables without superscript on the upper-left corner could be assumed to be expressed in the inertial frame. The illustration of the coordinate system could be found in Fig. \ref{fig:EoM}.

The input to the dynamical system is the GRF $\bm{u}_i \in \mathbb{R}^3$ at contact foot locations $\bm{p}_i^f \in \mathbb{R}^3$. The GRFs create the external wrench to the rigid body, where $i\in\{1,2,3,4\}$ is the index for the front left (FL), front right (FR), hind left (HL) and hind right (HR), respectively, as shown in Fig. \ref{fig:EoM}. The foot positions $\bm{p}_i^f$ relative to CoM are denoted as $\bm{r}_i = \bm{p}^f_i - \bm{p}$. Therefore, the net external wrench $\bm{\mathcal{F}}\in\mathbb{R}^6$ exerted on the body is:	
\begin{align}
	\bm{\mathcal{F}}=
	\begin{bmatrix}
	\bm{F}\\
	\bm{\tau}
	\end{bmatrix}=
	\sum_{i=1}^4
	\begin{bmatrix}
		\mathbb{I}\\
		\hat{\bm{r}}_i
	\end{bmatrix}\bm{u}_i,
\end{align}
where $\bm{F}$ and $\bm{\tau}$ are the total force and torque applied at the CoM; the hat map $\hat{(\cdot)}:\mathbb{R}^3 \rightarrow \mathfrak{so}(3)$ maps an element from $\mathbb{R}^3$ to the space of skew-symmetric matrices $\mathfrak{so}(3)$, which represents the cross-product under multiplication as $\hat{\bm{\alpha}}\bm{\beta}=\bm{\alpha}\times \bm{\beta}, \text{for all } \bm{\alpha},\bm{\beta}\in \mathbb{R}^3$. The inverse of the hat map is the vee map $(\cdot)^{\vee}:\mathfrak{so}(3) \rightarrow \mathbb{R}^3$. The full dynamics of the rigid body can be written as
\begin{equation}\label{eq:full_dynamics}
	\dot{\bm{x}}=
	\begin{bmatrix}
		\dot{\bm{p}}\\
		\ddot{\bm{p}}\\
		\dot{\bm{R}}\\
		^B\dot{\bm{\omega}}
	\end{bmatrix}
	=
	\begin{bmatrix}
		\dot{\bm{p}}\\
		\frac{1}{M}{\bm{F}}+\bm{a}_g\\
		\bm{R}\cdot\leftidx{^B}{\hat{\bm{\omega}}}\\
		^B\bm{I}^{-1}(\bm{R}^T{\bm{\tau}}-{^B\hat{\bm{\omega}}}{^B\bm{I}}{^B\bm{\omega}})
	\end{bmatrix},
\end{equation}
where $\bm{u}=[\bm{u}_1^T,\bm{u}_2^T,\bm{u}_3^T,\bm{u}_4^T]^T\in\mathbb{R}^{12}$ is the control vector; $M$ is the mass of the rigid body; $\bm{a}_g=[0,0,-g]^T$ is the gravitational acceleration vector; $^B\bm{I}\in\mathbb{R}^{3\times3}$ is the fixed moment of inertia tensor in the body frame $\{\bm{B}\}$. The inertia properties of the robot could be found in Table \ref{tab:parameters}.

\begin{table}
	\vspace{16px}		
	\caption{System Parameters of the Robot}
	\centering
	\begin{tabular}{ c c c }
		\hline
		Parameter 		& Value & Unit\\
		\hline
		M				& 5.5	& kg\\
		$I_{xx}$		& 0.026	& kg$\cdot$m$^2$\\
		$I_{yy}$		& 0.112	& kg$\cdot$m$^2$\\
		$I_{zz}$		& 0.075	& kg$\cdot$m$^2$\\
		Body length		& 0.3	& m\\
		Body width		& 0.2	& m\\
		Link length		& 0.14	& m\\
		\hline
	\end{tabular}
	\label{tab:parameters}
\end{table}

To develop a representation-free control approach, we decided to directly parameterize orientation using the rotation matrix. This completely avoids the singularities and complexities when using local coordinates such as Euler angles. It also avoids the ambiguities when using quaternions to represent attitude dynamics. As the quaternion double covers the special orthogonal group $SO(3)$,  the control design needs to switch between the local charts. 

The rotational dynamics of (\ref{eq:full_dynamics}) is nonlinear for it involves the rotation matrix $\bm{R}$, which evolves on the $SO(3)$ manifold. In Section \ref{sec:variation} we present a variation-based linearization scheme for linearizing the rotational dynamics.

\subsection{Variation-based Linearization}
\label{sec:variation}	
Although the nonlinear MPC (\ref{eq:nonlin_MPC}) could be solved to obtain the control input, the presence of local optimum resulting from the nonlinear dynamics complicates the solution process. Furthermore, convoluted nonlinear optimization does not lend itself well to embedded implementations. To meet the real-time constraint, we strive to formulate the MPC as a Quadratic Program (QP) that can be efficiently solved on embedded systems. A variation-based linearization scheme for the rotation matrix is proposed to linearize the nonlinear rotational dynamics. The error on the non-Euclidean $SO(3)$ manifold is approximated by the corresponding variation\cite{marsden1995introduction} with respect to the operating point. Then the variation dynamics is derived based on the system model (\ref{eq:full_dynamics}) in the manner of \cite{wu2015variation}. Recent work \cite{VBL_QP} achieved underactuated two-leg balancing on MIT Mini Cheetah using variational-based linearization on the $SO(3)$ manifold.

Assuming that the predicted variables are close to the operating point, the variation of the rotation matrix on $\mathfrak{so}(3)$ could be approximated by $\delta\bm{R}$ using the derivative of the error function on $SO$(3) as in \cite{lee2010geometric}. The variation $\delta\bm{R}\in\mathfrak{so}(3)$ is a local approximation of the displacement between two points on the $SO(3)$ manifold. The rotation matrix at the $k^{th}$ prediction step is approximated using the first-order Taylor expansion of matrix exponential map,
\begin{equation}\label{eq:R_k}
	\bm{R}_k\approx\bm{R}_{op}\text{exp}({\delta\bm{R}_k}) \approx\bm{R}_{op} (\mathbb{I}+\delta\bm{R}_k),
\end{equation}
where we use the commutativity of small rotations based on the assumption of $\delta\bm{R}$ being small.

The nonlinear dynamics of the rotation matrix is given as
\begin{equation}\label{eq: R_dot}
\dot{\bm{R}}=\bm{R}{^B\hat{\bm{\omega}}},
\end{equation}
where the first-order approximation of rotation matrix $\bm{R}_k$ is presented in (\ref{eq:R_k}). To get a linear approximation for $\dot{\bm{R}}$, we define the variation of angular velocity $\delta\bm{\omega}_k$ as
\begin{equation}\label{eq:delta_w_k}
{\delta \bm{\omega}_k} = {\bm{\omega}}_k - {\bm{R}_k^T \bm{R}_{op} \bm{\omega}_{op}},
\end{equation}
where the transport map $\bm{\omega}_{op}\rightarrow\bm{R}_k^T \bm{R}_{op} \bm{\omega}_{op}$ enables comparison between tangent vectors at different points. This procedure is required because the tangent vectors $\dot{\bm{R}}_k \in T_{R_k}SO(3)$ and $\dot{\bm{R}}_{op} \in T_{R_{op}}SO(3)$ lie in different tangent spaces and cannot be compared directly, where $ T_{R_{op}}SO(3)$ refers to the tangent space of $SO(3)$ at $\bm{R}_{op}$. Hence, the angular velocity $\bm{\omega}$ could be deducted from (\ref{eq:delta_w_k}) as
\begin{equation}\label{eq: w_hat}
	\begin{split}
		\bm{\omega}_k =& {\bm{R}_k^T \bm{R}_{op} \bm{\omega}_{op}} + {\delta \bm{\omega}_k}\\
		=& (\mathbb{I} + \delta \bm{R}_k)^T \bm{\omega}_{op} + \delta\bm{\omega}_k\\
		=& \bm{\omega}_{op} + \delta{\bm{\omega}}_k - \delta \bm{R}_k\bm{\omega}_{op},
	\end{split}
\end{equation}
where $\bm{R}_k$ is replaced by the expression in (\ref{eq:R_k}). The last step in (\ref{eq: w_hat}) is due to the fact that $\delta \bm{R}_k$ is a skew-symmetric matrix by construction. Applying the hat map $\hat{(\cdot)}$ to $\bm{\omega}_k$ and substituting (\ref{eq: w_hat}) into (\ref{eq: R_dot}) yields
\begin{equation}
	\begin{split}
		\dot{\bm{R}}_k &= \bm{R}_k \hat{\bm{\omega}}_k = \bm{R}_{op}(\mathbb{I}+\delta \bm{R}_k)(\hat{\bm{\omega}}_{op} + \widehat{\delta\bm{\omega}_k} - \widehat{\delta \bm{R}_k\bm{\omega}_{op}})\\
		&= \bm{R}_{op} \hat{\bm{\omega}}_{op} + \bm{R}_{op}\widehat{\delta\bm{\omega_k}} - \bm{R}_{op}\widehat{\delta \bm{R}_k\bm{\omega}_{op}} + \bm{R}_{op}\delta \bm{R}_k\hat{\bm{\omega}}_{op},
	\end{split}
\end{equation}
where the higher order variation terms are neglected. Using the properties of cross product in \cite{featherstone2014rigid} Table 2.1, the following equality
\begin{equation}
	\widehat{\delta \bm{R}_k\bm{\omega}_{op}} = \delta \bm{R}_k\hat{\bm{\omega}}_{op} - \hat{\bm{\omega}}_{op}\delta \bm{R}_k,
\end{equation}
is used to derive the expression of $\dot{\bm{R}}_k$
\begin{equation}\label{eq:linear_dR}
	\dot{\bm{R}}_k =  \bm{R}_{op} \hat{\bm{\omega}}_{op} + \bm{R}_{op}\hat{\bm{\omega}}_{op}\delta \bm{R}_k + \bm{R}_{op} \widehat{\delta \bm{\omega}_k}.
\end{equation}
The dynamics of angular velocity $\dot{\bm{\omega}}_k$ is linearized as
\begin{equation}\label{eq:I_dw}
	\begin{split}
		{^B\bm{I}}\dot{\bm{\omega}}_k = 
		& \bm{R}^T_{op}\bm{\tau}_{op}+\delta \bm{R}_k^T \bm{\tau}_{op} + \bm{R}_{op}^T \delta \bm{\tau}_k +  \\
		&-\hat{\bm{\omega}}_{op}{^B\bm{I}}\bm{\omega}_{op} -\hat{\delta \bm{\omega}_k} {^B\bm{I}} \bm{\omega}_{op} -\hat{\bm{\omega}}_{op} {^B\bm{I}} \delta \bm{\omega}_k,
	\end{split}
\end{equation}
in which $\delta\bm{\tau}_k$ is the variation of the net torque,
\begin{equation}\label{eq:delta_tau}
	\delta \bm{\tau}_k = (\sum_{i=1}^4 \hat{\bm{u}}_{i,op})\delta\bm{p}_k + (\sum_{i=1}^4 \hat{\bm{r}}_{i,op}\cdot\delta \bm{u}_{i,k}),
\end{equation}
where $\bm{u}_{op}$ is GRF applied at the current step, $\delta\bm{u}$ is the variation of GRF from $\bm{u}_{op}$. Note that the GRF applied at the next time step is $\bm{u}_{op}+\delta\bm{u}_1$.

The linearized dynamics will be used as affine dynamics as well as in the construction of objective function in the MPC formulation.

\subsection{Vectorization}
\label{sec:vectorization}	

After the dynamics of rotation matrix $\bm{R}$ and angular velocity $\bm{\omega}$ are linearized, the matrix variables in~\eqref{eq:linear_dR} and~\eqref{eq:I_dw} are still difficult to be formulated into the standard QP form. This section proposes a vectorization technique which uses Kronecker product \cite{graham2018kronecker} to transform matrix-matrix products into matrix-vector products.

Let us define a vector $\bm{\xi}\in \mathbb{R}^3$ be such that the skew-symmetric matrix $\hat{\bm{\xi}}=\delta \bm{R} \in \mathfrak{so}(3)$ is an element in the tangent space at the operating point. Let $\bm{N}\in\mathbb{R}^{9\times 3}$ be a constant matrix such that $vec(\hat{v})=\bm{N}v, \forall v \in\mathbb{R}^3$, where $vec(\cdot)$ is the vectorization function. Then $vec(\delta \bm{R})= \bm{N}\cdot \bm{\xi}$. The second and third terms of (\ref{eq:linear_dR}) could be vectorized as:
\begin{equation}\label{eq:vec_Rdot}
	\begin{split}
		& vec(\bm{R}_{op} \hat{\bm{\omega}}_{op}\delta \bm{R}_k) = (\mathbb{I}\otimes \bm{R}_{op}\hat{\bm{\omega}}_{op}) \bm{N} \bm{\xi}_k\\
		& vec{(\bm{R}_{op}\widehat{\delta\bm{\omega}_k})} = (\mathbb{I} \otimes \bm{R}_{op}) vec(\widehat{\delta\bm{\omega}_k}),
	\end{split}
\end{equation}
where $\otimes$ is the Kronecker tensor operator. To derive the expression for $vec(\widehat{\delta\bm{\omega}_k})$, one plugs (\ref{eq:R_k}) into (\ref{eq:delta_w_k})
\begin{equation}
	vec(\widehat{\delta\bm{\omega}_k}) = \bm{N}(\bm{\omega}_k-\bm{\omega}_{op} + \hat{\bm{\omega}}_{op}\bm{\xi}_k).
\end{equation}
The vectorized version of (\ref{eq:linear_dR}) is 
\begin{equation}\label{eq:vec_dR}
	vec(\dot{\bm{R}}_k) = \bm{C}_{\xi}^c + \bm{C}_{\xi} ^{\xi} \bm{\xi}_k + \bm{C}_{\xi}^{\omega}\bm{\omega}_k ,
\end{equation}
where the constants are defined as
\begin{equation}
	\begin{split}
		\bm{C}_{\xi}^c &= vec({\bm{R}_{op}}{\hat{\bm{\omega}} _{op}}) - (\mathbb{I} \otimes \bm{R}_{op})\bm{N}\bm{\omega}_{op}\\
		\bm{C}_{\xi}^{\xi} &= (\mathbb{I}\otimes \bm{R}_{op}{\hat{\bm{\omega}}_{op}}) \bm{N} - (\mathbb{I} \otimes \bm{R}_{op})\bm{N}\hat{\bm{\omega}}_{op}\\
		\bm{C}_{\xi}^{\omega} &= (\mathbb{I} \otimes \bm{R}_{op})\bm{N}.
	\end{split}
\end{equation}
The discrete orientation dynamics in terms of $\bm{\xi}$ is derived by propagating the rotation matrix using the forward Euler integration scheme,
\begin{equation}
	\bm{R}_{k+1} = \bm{R}_k + \dot{\bm{R}}_k dt,
\end{equation}
where $dt$ is the MPC sampling time. When the rotation matrix is approximated by the first-order expansion in (\ref{eq:R_k}), the above expression could be simplified to the following form,
\begin{equation}\label{eq:R_derive_1}
	\delta \bm{R}_{k+1} = \delta \bm{R}_k + \bm{R}_{op}^T \dot{\bm{R}}_k dt.
\end{equation}
Vectorizing (\ref{eq:R_derive_1}) gives
\begin{equation}
	\bm{N} \bm{\xi}_{k+1} = \bm{N} \bm{\xi}_{k} + dt (\mathbb{I}\otimes \bm{R}_{op}^T) vec(\dot{\bm{R}}_k).
\end{equation}
The discrete dynamics in $\bm{\xi}$ is obtained by putting in (\ref{eq:vec_dR}) and pre-multiply with $\bm{N}^*$, the left pseudo-inverse of $\bm{N}$
\begin{equation}\label{eq:eta_final}
	\bm{\xi}_{k+1} = \bm{\xi}_k + dt \bm{N}^{*} (\mathbb{I}\otimes \bm{R}_{op}^T) (\bm{C}_{\xi}^c + \bm{C}_{\xi}^{\xi}\bm{\xi}_k + \bm{C}_{\xi}^{\omega}\bm{\omega}_k).
\end{equation}

The angular velocity dynamics in (\ref{eq:I_dw}) is also vectorized. The second term in (\ref{eq:I_dw}) is $\delta \bm{R}_k^T \bm{\tau}_{op} = (\mathbb{I}\otimes\bm{\tau}_{op}^T)\bm{N}\bm{\xi}_k$ and $\delta \bm{\tau}_k$ is defined in (\ref{eq:delta_tau}). The last two terms could be further derived,
\begin{equation}
	\begin{split}
		-& \hat{\delta \bm{\omega}_k} {^B\bm{I}} \bm{\omega}_{op} -\hat{\bm{\omega}}_{op} {^B\bm{I}} \delta \bm{\omega}_k \\
		&= (\widehat{{^B\bm{I}} \bm{\omega}_{op}} - \hat{\bm{\omega}}_{op} {^B\bm{I}}) \delta \bm{\omega}\\
		&= (\widehat{{^B\bm{I}} \bm{\omega}_{op}} - \hat{\bm{\omega}}_{op} {^B\bm{I}})(\bm{\omega}_k-\bm{\omega}_{op}-\hat{\bm{\omega}}_{op}\bm{\xi}_k).
	\end{split}
\end{equation}
Assembling these expressions into (\ref{eq:I_dw}) and rearranging the corresponding terms gives the vectorization of $^B\bm{I}\dot{\bm{\omega}}_k$
\begin{equation}\label{eq:wdot_final}
	^B\bm{I}\dot{\bm{\omega}}_k = \bm{C}_{\dot{\omega}} + \bm{C}_{\dot{\omega}}^{\delta p}\bm{p}_k + \bm{C}_{\dot{\omega}}^{\xi}\bm{\xi}_k + \bm{C}_{\dot{\omega}}^{\omega}\bm{\omega}_k + \bm{C}_{\dot{\omega}}^{\delta u}\delta \bm{u}_k,
\end{equation}
where
\begin{equation}\label{eq:wdot_coeff}
	\begin{split}
		\bm{C}_{\dot{\omega}}^c = & -\hat{\bm{\omega}}_{op}{^B\bm{I}}\bm{\omega}_{op} + \bm{R}^{T}_{op}\bm{\tau}_{op} - (\widehat{^B\bm{I} \bm{\omega}_{op}} - \hat{\bm{\omega}}_{op}{^B\bm{I}})\bm{\omega}_{op}\\
		& - \bm{R}_{op}^T(\sum\hat{\bm{u}}_{op})\bm{p}_{op}\\
		\bm{C}_{\dot{\omega}}^{\delta p} = & ~\bm{R}_{op}^T(\sum_i\hat{\bm{u}}_{op}^i)\\
		\bm{C}_{\dot{\omega}}^{\xi} = & ~(\mathbb{I}\otimes\bm{\tau}_{op}^T)\bm{N} - (\widehat{^B\bm{I} \bm{\omega}_{op}} - \hat{\bm{\omega}}_{op}{^B\bm{I}})\hat{\bm{\omega}}_{op} \\
		\bm{C}_{\dot{\omega}}^{\omega} = & ~\widehat{^B\bm{I} \bm{\omega}_{op}} - \hat{\bm{\omega}}_{op}{^B\bm{I}}\\
		\bm{C}_{\dot{\omega}}^{\delta u} = & ~\bm{R}_{op}^T[\hat{\bm{r}}_{op}^1,\hat{\bm{r}}_{op}^2,\hat{\bm{r}}_{op}^3,\hat{\bm{r}}_{op}^4].\\
	\end{split}
\end{equation}

The discrete dynamics of $\bm{\omega}$ is propagated using forward Euler scheme $\bm{\omega}_{k+1} = \bm{\omega}_k + dt \dot{\bm{\omega}}_k$.

\subsection{Discrete-time Affine Dynamics}
\label{sec:discrete_dyn}

The single rigid body model introduced in Section \ref{sec:nonlin_dyn} has nonlinear dynamics in $\bm{R}$ and $\bm{\omega}$. Hence, a variation-based linearization scheme is proposed in Section \ref{sec:variation} to linearize the nonlinear dynamics. Section \ref{sec:vectorization} presents a vectorization method to reformulate the matrix variables into the vector form. Based on the aforementioned procedures, the new set of state and control vectors are defined as,
\begin{equation}\label{eq:new_state_control}
	\begin{split}
		\bm{x}_k & :=[\bm{p}_k^T~~\dot{\bm{p}}_k^T~~\bm{\xi}_k^T~~^B\bm{\omega}_k^T]^T\in\mathbb{R}^{12} \\
		\delta\bm{u}_k & :=[\delta \bm{u}_{1,k}^T~~\delta \bm{u}_{2,k}^T~~\delta \bm{u}_{3,k}^T~~\delta \bm{u}_{4,k}^T]^T \in \mathbb{R}^{12},
	\end{split}
\end{equation}
where the new control input $\delta \bm{u}_i\in\mathbb{R}^3$ is the variation of GRF from the operating point $\bm{u}_{i,op}$ for the $i^{th}$ leg. 

By assembling the corresponding terms from (\ref{eq:eta_final}), (\ref{eq:wdot_final}) into matrix form, the discrete-time affine dynamics could be expressed in the state-space form:
\begin{equation}\label{eq:discrete_dyn}
	\bm{x}_{t+k+1|t} = \bm{A}|_{op}\cdot\bm{x}_{t+k|t} + \bm{B}|_{op}\cdot\delta\bm{u}_{t+k|t}+\bm{d}|_{op},
\end{equation}
where $\bm{A}|_{op}\in\mathbb{R}^{n\times n},\bm{B}|_{op}\in\mathbb{R}^{n\times m}$, and $\bm{d}|_{op}\in\mathbb{R}^n$ are matrices constructed by the measurements at the operating point. Therefore, nonlinear dynamics have been linearized about the operating point to result in a locally-valid linear time-varying (LTV) system. This system can be stabilized to track reference trajectories.

The discrete-time affine dynamics are imposed as equality constraint as in (\ref{eq:nonlin_MPC}b).

\subsection{Cost Function}
\label{sec:cost_fcn}
The cost function in the nonlinear MPC formulation (\ref{eq:nonlin_MPC}) includes both terminal and stage costs. In this work, the cost is set as a quadratic function that penalizes deviation from the reference trajectories. The stage cost is
\begin{equation}\label{eq:stage_cost}
	\ell(\bm{x}_k,\bm{u}_k) =||\bm{x}_k-\bm{x}_{d,k}||_{\bm{Q}_x}^2 + ||\bm{u}_{k}-\bm{u}_{d,k}||_{\bm{R}_u}^2,
\end{equation}
where $||\bm{x}||_{\bm{Q}}^2$ is a shorthand notation of the matrix norm $\bm{x}^T\bm{Q}\bm{x}$ where $\bm{Q}$ is a positive definite matrix; $\bm{x}_{d,k}$ and $\bm{u}_{d,k}$ are the desired state and control at the $k^{th}$ prediction step; $\bm{Q}_x$ and $\bm{R}_u$ are the block diagonal positive definite gain matrices for state and control, respectively. The first term in (\ref{eq:stage_cost}) consisting of the error functions of the state vector could be decomposed as:
\begin{equation}\label{eq:state_deviation}
	||\bm{x}_k-\bm{x}_{k,d}||_{\bm{Q}}^2 = ||\bm{e}_{p_k}||_{\bm{Q}_p}^2 + ||\bm{e}_{\dot p_k}||_{\bm{Q}_{\dot p}}^2 + ||\bm{e}_{R_k}||_{\bm{Q}_R}^2 + ||\bm{e}_{\omega_k}||_{\bm{Q}_{\omega}}^2,
\end{equation}
where $\bm{Q}_p,\bm{Q}_{\dot{p}},\bm{Q}_{R},\bm{Q}_{\omega}$ are diagonal positive definite weighting matrices; $\bm{e}_{p_k},\bm{e}_{\dot{p}_k}$ are the error terms for deviations from the corresponding reference trajectories $\bm{p}_k^d, \dot{\bm{p}}_k^d$ constructed from the user input. The error function for the rotation matrix and angular velocity are given by \cite{bullo2004geometric} as
\begin{equation}\label{eq:error_R}
	\bm{e}_{R_k} = \text{log}(\bm{R}_{d,k}^T \cdot \bm{R}_k)^{\vee}
\end{equation}
\begin{equation}\label{eq:error_w}
	\bm{e}_{\omega_k} = {\bm{\omega}}_k - \bm{R}_k^T  \bm{R}_{d,k} \bm{\omega}_{d,k},
\end{equation}
where $\bm{R}_{d,k}$ and $\bm{\omega}_{d,k}$ are the desired rotation matrix and angular velocity trajectories. The terminal cost function is similarly defined.

In the stage cost expression (\ref{eq:state_deviation}), all the error terms are in linear form of the state and control vectors except the error of orientation $\bm{e}_{R_k}$, which involves matrix logarithm map as shown in (\ref{eq:error_R}). A linear approximation of the nonlinear error term on the rotation matrix is used. Taking the hat map on (\ref{eq:error_R}) and applying the matrix exponential map give
\begin{equation}\label{eq:cost_derive}
	\text{exp}(\hat{\bm{e}}_{R_k})=\bm{R}_{d,k}^T\bm{R}_k\approx \bm{R}_{d,k}^T \bm{R}_{op}\text{exp}(\hat{\bm{\xi}}_k),
\end{equation}
where the same approximation is made here as in (\ref{eq:R_k}). Taking the matrix logarithm of (\ref{eq:cost_derive}) and then applying the vee map give the approximate error term on $\bm{R}_k$,
\begin{equation}\label{eq:orientation_error}
	\bm{e}_{R_k} = \text{log}(\bm{R}_{d,k}^T \cdot \bm{R}_{op})^{\vee} + \bm{\xi}_k.
\end{equation}

The error function defined in (\ref{eq:orientation_error}) is in linear form of the state variable $\bm{\xi}_k$ since both $\bm{R}_{d,k}^T$ and $\bm{R}_{op}$ are known matrices. The error function could be interpreted as the sum of (a) the geodesic between $\bm{R}_{op}$ and $\bm{R}_{d,k}$ and (b) the vector $\bm{\xi}_k$ which lies in the tangent space at $\bm{R}_{op}$. The orientation error function $\Psi$ on $\bm{R}$ could therefore be defined as,
\begin{equation}
	\Psi(\bm{\xi}_k) = \bm{e}_{R_k}^T \bm{Q}_R \bm{e}_{R_k} = ||\bm{e}_{R_k}||_{\bm{Q}_R}^2,
\end{equation}
which is in quadratic form of $\bm{\xi}_k$. Given that the weighting matrix $\bm{Q}_R$ is positive definite, the orientation error function $\Psi$ is positive definite.

The terminal cost $\ell_T$ is similarly defined as (\ref{eq:state_deviation}) with terminal gains. The cost of control is constructed as
\begin{equation}
	||\bm{u}_{k}-\bm{u}_{d,k}||_{\bm{R}_u}^2 = ||\bm{u}_{op} + \delta\bm{u}_k - \bm{u}_{d,k}||_{\bm{R}_u}^2,
\end{equation}
where $\delta\bm{u}_k$ is the $k^{th}$ predicted variation from the operating point $\bm{u}_{op}$.

\subsection{Force Constraints}
\label{sec:ineq}

The force constraints are imposed to ensure that the solved GRF are physically feasible. When the foot is in contact with the ground, the normal force should be non-negative and the tangent forces should lie within the friction cone, which is prescribed as
\begin{equation}
	\{\bm{u}_i | \bm{u}^n_i \geq  0, ||\bm{u}_i^t||_2 \leq \mu |u_i^n|\},
\end{equation}
where $\mu$ is the coefficient of friction; superscript $(\cdot)^t$ and $(\cdot)^n$ indicate tangential and normal force components, respectively. $||\cdot||_2$ is the 2-norm and $|\cdot|$ takes the absolute value of a scalar.

Since the friction cone constraint is a second-order cone constraint, it is not admissible to the QP formulation with linear constraints. Instead, the conservative friction pyramid\cite{trinkle1997dynamic} is used as an approximation of the friction cone. In addition, the normal force is bounded to ensure that the commanded torque does not exceed the actuator limits. The feasible control set $\mathbb{U}$ in (\ref{eq:nonlin_MPC}) is defined as:
\begin{equation}\label{eq:force_constraint}
	\begin{split}
		\mathbb{U}_i:=\{\delta\bm{u}_i ~|~ & |u_{i,op}^{x/y}+\delta u_{i,k}^{x/y}| \leq \mu |u_{i,op}^z + \delta u_{i,k}^z|, \\
		~ & u_{i,k}^{z,min} \leq u_{i,op}^z + \delta u_{i,k}^z \leq u_{i,k}^{z,max},\\
		~ & u_{i,k}^{z,min}\geq 0\},
	\end{split}
\end{equation}
where the $z$ axis is aligned with the normal vector of the ground and $x, y$ are two axes orthogonal to each other that lie in the tangent plane at the contact point; $u_{i,k}^{z,min}$ and $u_{i,k}^{z,max}$ are the minimum and maximum normal forces at the $k^{th}$ predicted step for leg $i$. If leg $i$ was in swing phase, then the value for both lower and upper bounds are set to zero so that the swing leg controller takes over and guides the foot towards the next foothold position. It could be observed that (\ref{eq:force_constraint}) denotes an intersection of a finite set of closed halfspaces in $\mathbb{R}^3$. Hence, $\mathbb{U}_i$ is a polyhedron. Similarly, the feasible force set $\mathbb{U}$ is also polyhedral.

\subsection{Quadratic Program Formulation}
\label{sec:qp}
Given the convex quadratic cost function from Section \ref{sec:cost_fcn}, the affine dyanmics from Section \ref{sec:discrete_dyn} and linear force constraints from Section \ref{sec:ineq}, the general nonlinear MPC problem (\ref{eq:nonlin_MPC}) could be be reformulated as a Quadratic Program (QP),
\begin{subequations}\label{eq:lin_MPC}
	\begin{align}
		\text{min.}& ~~~~~~ \gamma^{N}\ell_T(\bm{x}_{t+N|t}) + \sum_{k=0}^{N-1}\gamma^{k}\ell(\bm{x}_{t+k|t},\delta\bm{u}_{t+k|t})\\
		\text{s.t.}& ~~~~~~ \bm{x}_{t+k+1|t} = \bm{A}\bm{x}_{t+k|t}+\bm{B}\delta\bm{u}_{t+k|t}+\bm{d}\\
		& ~~~~~~ \delta\bm{u}_{t+k|t} \in \mathbb{U}_k, k=0,\cdots,N-1\\
		& ~~~~~~ \bm{x}_{t|t} = \bm{x}(t)=\bm{x}_{op},
	\end{align}
\end{subequations}
where the cost function is defined in (\ref{eq:stage_cost}); the decay rate factor $\gamma\in(0,1]$ discounts cost further from the current moment. The definition of the affine dynamics could be found in (\ref{eq:discrete_dyn}) and the force constraint is expressed in (\ref{eq:force_constraint}). Note that we lifted the explicit constraints on the state vectors but instead relied on the cost function for the state regulation. The QP in (\ref{eq:lin_MPC}) could be rewritten in a more compact form. Following the formulation in \cite{5153127}, the new optimization variable $\bm{z}$ is constructed as
\begin{equation}
	\bm{z}=[\delta\bm{u}_0^T, \bm{x}_1^T,\cdots,\delta\bm{u}_{N-1}^T,\bm{x}_N^T]^T\in\mathbb{R}^{24N},
\end{equation}
such that (\ref{eq:lin_MPC}) could be transcribed into the standard QP form\cite{boyd2004convex},
\begin{equation}
	\label{eq:QP}
	\begin{split}
		\text{minimize}& ~~~~~~ \frac{1}{2}\bm{z}^T \bm{P} \bm{z} + \bm{c}^T\bm{z}\\
		\text{subject to}& ~~~~~~ \bm{A}_{ineq}\cdot\bm{z} \leq \bm{b}_{ineq}\\
		& ~~~~~~ \bm{A}_{eq}\cdot\bm{z}=\bm{b}_{eq},
	\end{split}
\end{equation}
where $\bm{P}\in\mathbb{R}^{N(n+m)}$ is a symmetric positive definite matrix assembled from the gain matrices $\bm{Q}_x,\bm{R}_u$; the inequality constraint $\bm{A}_{ineq}\cdot\bm{z} \leq \bm{b}_{ineq}$ imposes the force constraints; the equality constraint $\bm{A}_{eq}\cdot\bm{z}=\bm{b}_{eq}$ respects the linear dynamics.

\section{Numerical Results}
\label{sec:Simulation}
This section first defines a function that quantifies the closeness to singularity for Euler angles in the numerical sense. Then we present the simulation results of RF-MPC stabilizing various periodic gaits and an aperiodic 3D acrobatic maneuver. Furthermore, RF-MPC is compared with the MPC that uses Euler angles as orientation representation (EA-MPC) in the acrobatic maneuver. The resulting QPs from the MPC formulation in all of the simulations are solved using MATLAB $quadprog$. Gain values and gait pattern parameters for the following simulations could be found in Table \ref{tab:gains}.

\subsection{Singularity in Euler Angles}
\label{sec:singularity_metric}
Orientation representations include Axis-angle, Euler angles and quaternion\cite{craig2009introduction}. In this work, we compare the proposed RF-MPC with EA-MPC, which is based on the convex MPC \cite{di2018dynamic} used on MIT Mini Cheetah, with parameters from \cite{miniGithub}.

Here, we define a function to quantify the distance to the singularity of Euler angles. Such function helps us to identify singularity and its neighborhood. Assuming EA-MPC adopts the Z-Y-X sequence in body frame $\{\bm{B}\}$, which is equivalent to the X-Y-Z sequence in stationary inertial frame $\{\bm{S}\}$. The Euler angles $\bm{\Theta}=[\phi~\theta~\psi]^T$, where $\phi$ is the roll, $\theta$ is the pitch, and $\psi$ is the yaw. The attitude of frame $\{\bm{B}\}$ is expressed by a sequence of rotations in frame $\{\bm{S}\}$ as
\begin{equation}
\bm{R} = \bm{R}_z(\psi)\bm{R}_y(\theta)\bm{R}_x(\phi),
\end{equation}
where $\bm{R}_x(\phi)$ means a positive rotation of angle $\phi$ around the $x$-axis of frame $\{\bm{S}\}$. 

We define $\mathcal{T}_{\bm{\Theta}}:\mathbb{R}^3\rightarrow \mathbb{R}^{3\times3}$ to be the matrix that converts $\dot{\bm{\Theta}}$ to the angular velocity expressed in $\{\bm{S}\}$ as
\begin{equation}\label{eq:dEA2om}
\bm{\omega}=\mathcal{T}_{\bm{\Theta}}\cdot\dot{\bm{\Theta}}=
\begin{bmatrix}
\cos(\theta)\cos(\psi) & -\sin(\psi) & 0\\
\cos(\theta)\sin(\psi) & \cos(\psi)  & 0\\
-\sin(\theta) & 0 & 1
\end{bmatrix}
\dot{\bm{\Theta}}.
\end{equation}
The matrix $\mathcal{T}_{\bm{\Theta}}$ in equation (\ref{eq:dEA2om}) is invertible when $\theta\neq\pm \frac{\pi}{2}$, and $\dot{\bm{\Theta}}$ could be calculated using the following equation
\begin{equation}
\dot{\bm{\Theta}} = \begin{bmatrix}
\cos(\psi)/\cos(\theta) & \sin(\psi)/\cos(\theta) & 0\\
-\sin(\psi) & \cos(\psi) & 0\\
\cos(\psi)\tan(\theta) & \sin(\psi) \tan(\theta) & 1
\end{bmatrix}\bm{\omega}.
\end{equation}

In this work, we use the function
\begin{equation}
\kappa^{-1}(\mathcal{T}_{\bm{\Theta}})\in(0,1]
\end{equation}
to quantify singularity in Euler angles, where $\kappa(\cdot)$ calculates the condition number of a matrix. When the robot approaches singular poses, the condition number $\kappa(\mathcal{T}_{\bm{\Theta}})$ increases rapidly, and its inverse $\kappa^{-1}(\mathcal{T}_{\bm{\Theta}})$ tends to 0. 

\begin{figure}
	\centering
	\resizebox{1\linewidth}{!}{\includegraphics{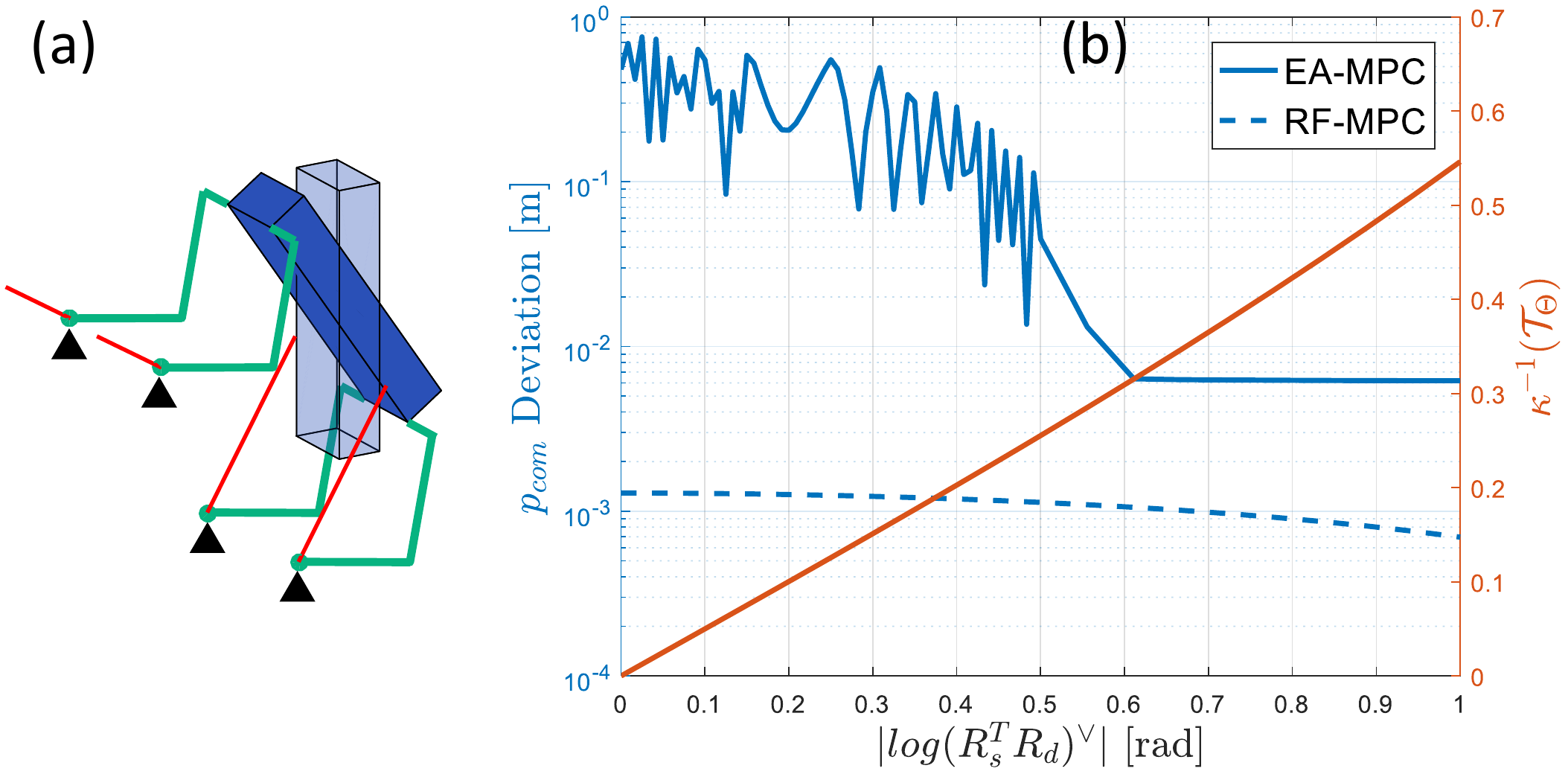}}
	\caption{The pose control simulation result of a investigation on the singularity of Euler angles. (a) The schematics of the pose control, where the shaded box represents the singular pose; the solid box represents the commanded pose; the red lines represent the GRF, with the force constraints (\ref{eq:force_constraint}) lifted. (b) The CoM position deviations (log scale) after a 0.5 s simulation of both RF-MPC and EA-MPC are plotted against $|log(\bm{R}_s^T\bm{R}_d)^{\vee}|$.}
	\label{fig:EA_GOT}
\end{figure}

Here, a pose control simulation is conducted to investigate the singularity of Euler angles. As shown in Fig. \ref{fig:EA_GOT}(a), the singular pose $\bm{R}_s$ is shown as the shadowed box; the desired pose $\bm{R}_d$ is shown as the solid box. All feet of the robot are assumed to be fix in this simulation so that the force constraints (\ref{eq:force_constraint}) could be lifted to focus on the effect of singularity. The desired poses are varied from the singular pose to the pose rotated 1 rad around the +y axis. A 0.5 s simulation is conducted in each desired pose and the CoM deviation at the end of the simulation is plotted for both RF-MPC and EA-MPC. As could be observed in Fig. \ref{fig:EA_GOT}(b), while RF-MPC remains stable, EA-MPC is significantly affected by singularity once $|log(\bm{R}_s^T\bm{R}_d)^{\vee}|<0.3$ rad, which corresponds to $\kappa^{-1}(\mathcal{T}_{\bm{\Theta}})<0.15$.

\subsection{Walking Trot}
\label{sec:sim_trot}
\begin{figure}
	\centering
	\resizebox{1\linewidth}{!}{\includegraphics{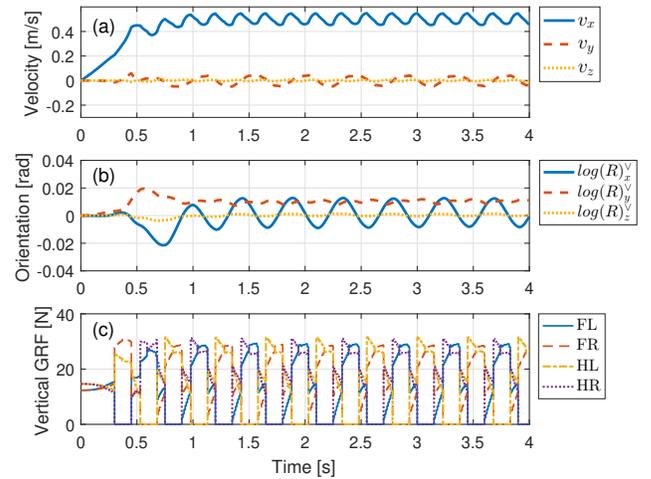}}
	\caption{Simulation data of walking trot where the robot starts from static pose and accelerates in the $x$-direction (a) CoM velocity; the robot accelerates at 0.5 m/s$^2$ and reaches the desired velocity of 0.5 m/s in the $x$-direction. (b) Orientation in terms of the log map of the rotation matrix (c) Vertical ground reaction forces of four legs.}
	\label{fig:sim_trot}
\end{figure}

The data of a walking trot simulation is shown in Fig. \ref{fig:sim_trot}. The robot starts from a stationary pose and accelerates in the $x$-direction until it reaches the final velocity of 0.5 m/s. As could be seen in Fig. \ref{fig:sim_trot} (a), the velocity in the $x$-direction reaches 0.5 m/s and the velocity deviation for all directions is within $\pm0.1~\text{m/s}$. Fig. \ref{fig:sim_trot} (b) shows that the orientation deviation in all directions is bounded within $\pm0.02~\text{rad}$. The vertical GRFs of all four legs are shown in Fig. \ref{fig:sim_trot} (c). Further details about the generation of the reference trajectory for trotting could be found in Section \ref{sec:gen_ref_traj}.

The simulation is setup such that at each sampling time, the control input is applied to the original nonlinear model (\ref{eq:full_dynamics}) simulated using MATLAB $ode45$ to integrate the dynamics. The gait pattern in the walking trot simulation is executed using a time-based schedule.

\subsection{Bounding}
\label{sec:sim_bound}
To demonstrate the capability of RF-MPC to stabilize dynamic motions with large body attitude oscillation, the bounding simulation is presented. The bounding motion involves an aerial phase when all four feet lose contact with the ground. To stabilize the bounding motion, the reference trajectory is designed based on the impulse-scaling principle \cite{park2017high} to preserve the periodicity of the gait. Details about the generation of reference trajectory could be found in Section \ref{sec:ref_gen_bound}. Here, we kindly note that an elaborate reference trajectory is optional for RF-MPC to stabilize bounding. While a trivial reference such as that used for trotting also works, the reference trajectory presented in Section Section \ref{sec:ref_gen_bound} enables bounding motions with longer aerial phase. Similar to the walking trot, the robot is commanded to start from the static pose and accelerates at 1.0 m/s$^2$ to reach the final velocity of 2.5 m/s.

\begin{figure}
	\centering
	\resizebox{1\linewidth}{!}{\includegraphics{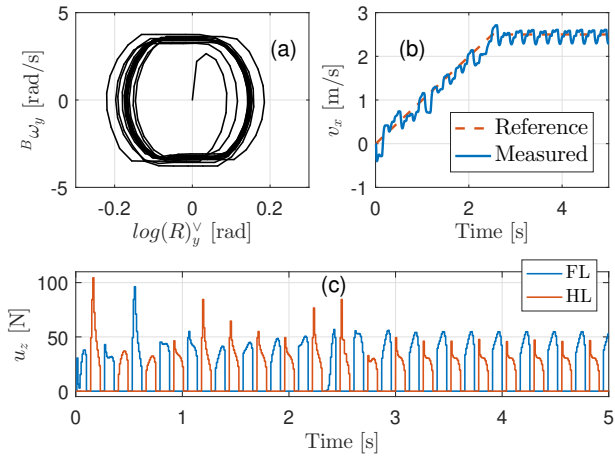}}
	\caption{Simulation data of bounding. (a) Phase portrait of body angle and angular velocity in the $y$-axis. (b) Velocity tracking performance in the $x$-direction. (c) Vertical GRFs of FL and HL legs.}
	\label{fig:sim_bound}
\end{figure}

Fig. \ref{fig:sim_bound} (a) presents the phase portrait of angle and angular velocity along the $y$-axis. It shows that the motion converges to a periodic orbit. The velocity tracking performance shown in Fig. \ref{fig:sim_bound} (b) demonstrates that the MPC controller is capable of stabilizing bounding velocity up to 2.5 m/s. The vertical GRF profiles of legs FL and HL are shown in Fig. \ref{fig:sim_bound} (c).

\subsection{Aperiodic Complex Dynamic Maneuver}
\label{sec:sim_acrobatic}
A complex acrobatic dynamic maneuver is presented in this section to demonstrate that RF-MPC is capable of controlling aperiodic dynamic motions that involves orientations that correspond to singularities in Euler angle representation. RF-MPC is benchmarked with an MPC controller with Euler angles for its orientation representation. In addition, initial condition is perturbed to investigate the robustness of RF-MPC.

\begin{figure}
	\centering
	\resizebox{0.9\linewidth}{!}{\includegraphics{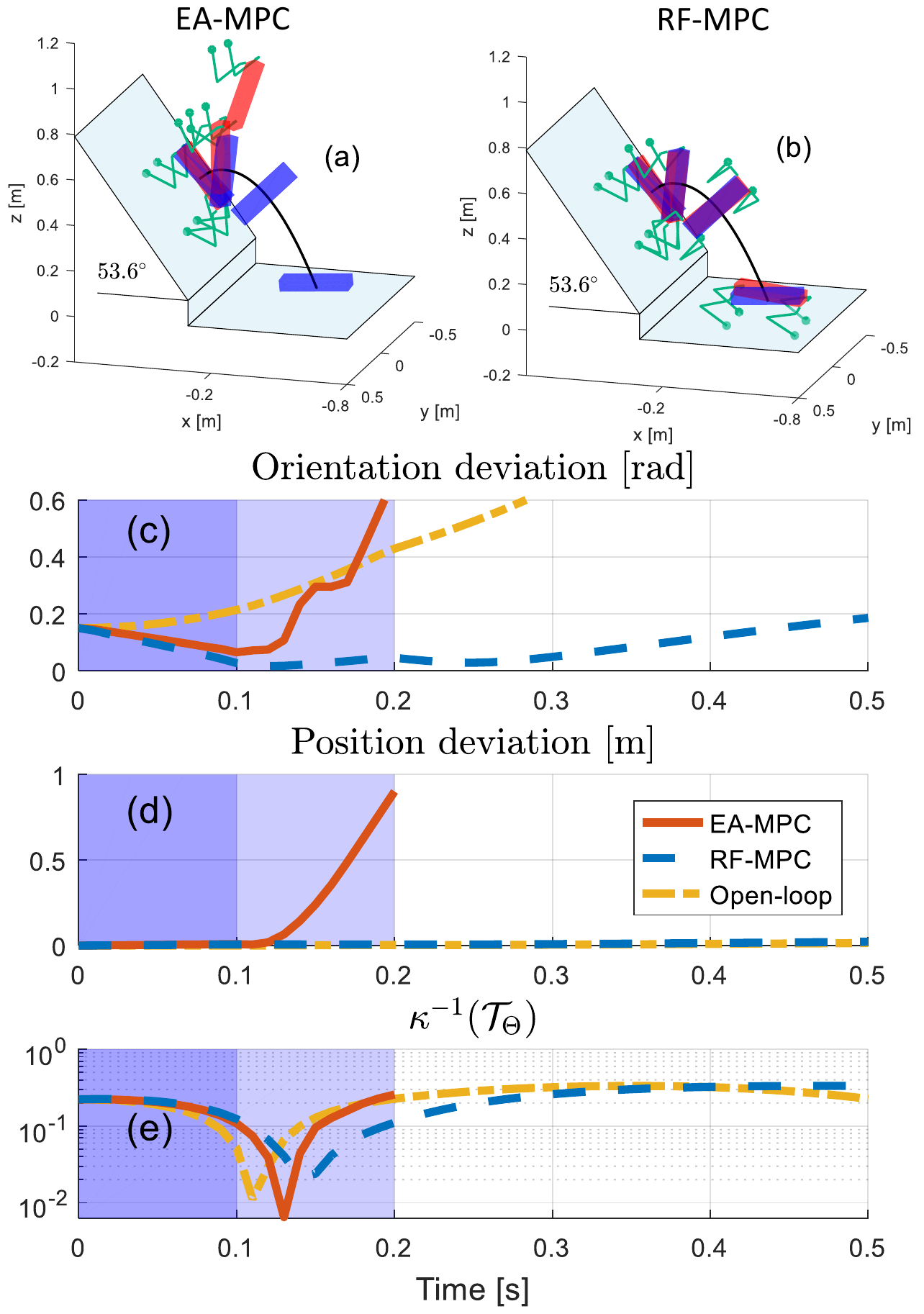}}
	\caption{Simulation results of a complex aperiodic 3D maneuver where the robot performs a twist jump off an inclined surface. The reference poses are shown in blue and the poses controlled by MPC are shown in red; the reference CoM trajectory is shown in black. The deep-shaded area is when four legs are in contact with the surface; the light-shaded area is when only hind legs are in contact and the non-shaded area is when the robot is in aerial phase. (a) The EA-MPC becomes unstable when the robot is close to the singular pose. (b) RF-MPC could track the reference motion that involves singular poses. (c) The orientation deviation $|\bm{e}_R|$ of open-loop control, RF-MPC, and the EA-MPC. (d) The CoM position deviation. (e) The function to quantify the distance to singularity $\kappa^{-1}(\mathcal{T}_{\bm{\Theta}})$.}
	\label{fig:twist_jump}
\end{figure}

Fig. \ref{fig:twist_jump} (a) shows the reference trajectory of the acrobatic motion, which is a backflip with a twist. The reference CoM trajectory is colored black and the reference poses are shown in blue. The robot initially stands on a slope with the slope angle of $45^\circ$, with the front of the robot facing upwards and body parallel to the slope. The stance phase of this acrobatic jump consists of 0.1 s of all four feet in contact, followed by 0.1 s of only the hind feet pair in contact with the slope. After the stance phase, all four feet are airborne and the robot enters the aerial phase for 0.3 s before landing. The landing direction of the robot is facing away from the slope. The feed-forward GRF and the dynamically-feasible reference trajectory are generated by solving an off-line TO problem, where the slope is set to be $45^\circ$. Further details about the generation of the reference trajectory is provided in Section \ref{sec:gen_ref_traj}.

As could be observed from Fig. \ref{fig:twist_jump} (a), when the robot approaches the singularity, EA-MPC becomes unstable and exerted large vertical force that pushes the robot away from the reference trajectory. As shown in Fig. \ref{fig:twist_jump} (c) (d), the body orientation and CoM position eventually diverges from the reference trajectory after the robot encounters singularity, which is visualized in Fig. \ref{fig:twist_jump} (e). In comparison, Fig. \ref{fig:twist_jump} (b) shows that RF-MPC could successfully stabilize the backflip motion that involves singularity. Here, we would like to point out in \ref{fig:twist_jump} (e) that the robot actually passes through singularity when the hind legs are in contact as indicated by the duration when $\kappa^{-1}(\mathcal{T}_{\bm{\Theta}})<10^{-1}$, in the light shaded area. Fig. \ref{fig:twist_jump} (c) (d) display that the orientation and CoM position deviation are kept small during the motion. The CoM position and orientation deviations are defined as 
\begin{equation}
\begin{aligned}
|\bm{e}_p| &= |\bm{p}(t)-\bm{p}_d(t)|\\
|\bm{e}_R| &=|log(\bm{R}_d^T(t)\bm{R}(t))^{\vee}|,
\end{aligned}
\end{equation}
where $\bm{p}_d$ and $\bm{R}_d$ are the reference CoM position and body orientation, respectively.

To demonstrate the robustness of RF-MPC, the slope angle for the simulated cases is changed from $45^\circ$ to $53.6^\circ$. Since the body of the robot is parallel to the slope at the beginning of the jump, the initial orientation of the robot is also perturbed. As could be observed from Fig. \ref{fig:twist_jump} (c), the backflip could be executed and stabilized by RF-MPC. In contrast, an open-loop simulation shows that in the absence of feedback control, the orientation of the robot quickly deviates from the reference trajectory due to the initial condition perturbation.

\subsection{Comparison of Linearization Schemes}
\label{sec:compare_linearizations}
One of the crucial decisions we made in the proposed RF-MPC is to linearize the dynamics about the operating point. The choice is made because RF-MPC represents orientation using the rotation matrix, which presumes $SO(3)$ structure. Such a structure loses its validity when the predicted states are far away from the point where the linearization is performed upon. Nevertheless, a reasonable alternative is linearizing around the reference trajectory, which is a widely used technique in robotics. To investigate which linearization scheme provides more robust behavior, this section presents a simulation case study that compares MPC linearized around the reference trajectory (scheme 1) with MPC linearized around the operating point (scheme 2). 

Scheme 1 linearizes dynamics around the reference trajectory, which includes the desired state $\{\bm{x}^d_{t+k|t}\}$ and control $\{\bm{u}^d_{t+k|t}\}$ within the prediction horizon, where $k=0,\cdots,N-1$ and $N$ is the prediction horizon. Hence, $\bm{A}_k$ and $\bm{B}_k$ are matrices for a Linear Time-Varying (LTV) system, parametrized by the reference trajectory within the prediction horizon. Scheme 2 linearizes dynamics around the operating point, which involves current state $\bm{x}_{op}$ and control $\bm{u}_{op}$, as defined in Section \ref{sec:MPC_formulation}. Constant matrices $\bm{A}|_{op}$ and $\bm{B}|_{op}$ are used to propagate the state through the prediction horizon using a Linear Time-Invariant (LTI) system.

The robustness of these two linearization schemes are qualitatively compared by examining how much external disturbance they can handle. The simulation is setup such that robot is bounding at a constant speed of 1.0 m/s in the $+x$ direction. The disturbance with a maximum force of 27 $N$ is applied on the robot in the $+y$ direction, causing it to deviate from the reference trajectory. The reference trajectory, controller gain, gait timing and disturbance are same for both schemes, with only the linearization scheme being different.

\begin{figure}
	\centering
	\resizebox{0.9\linewidth}{!}{\includegraphics{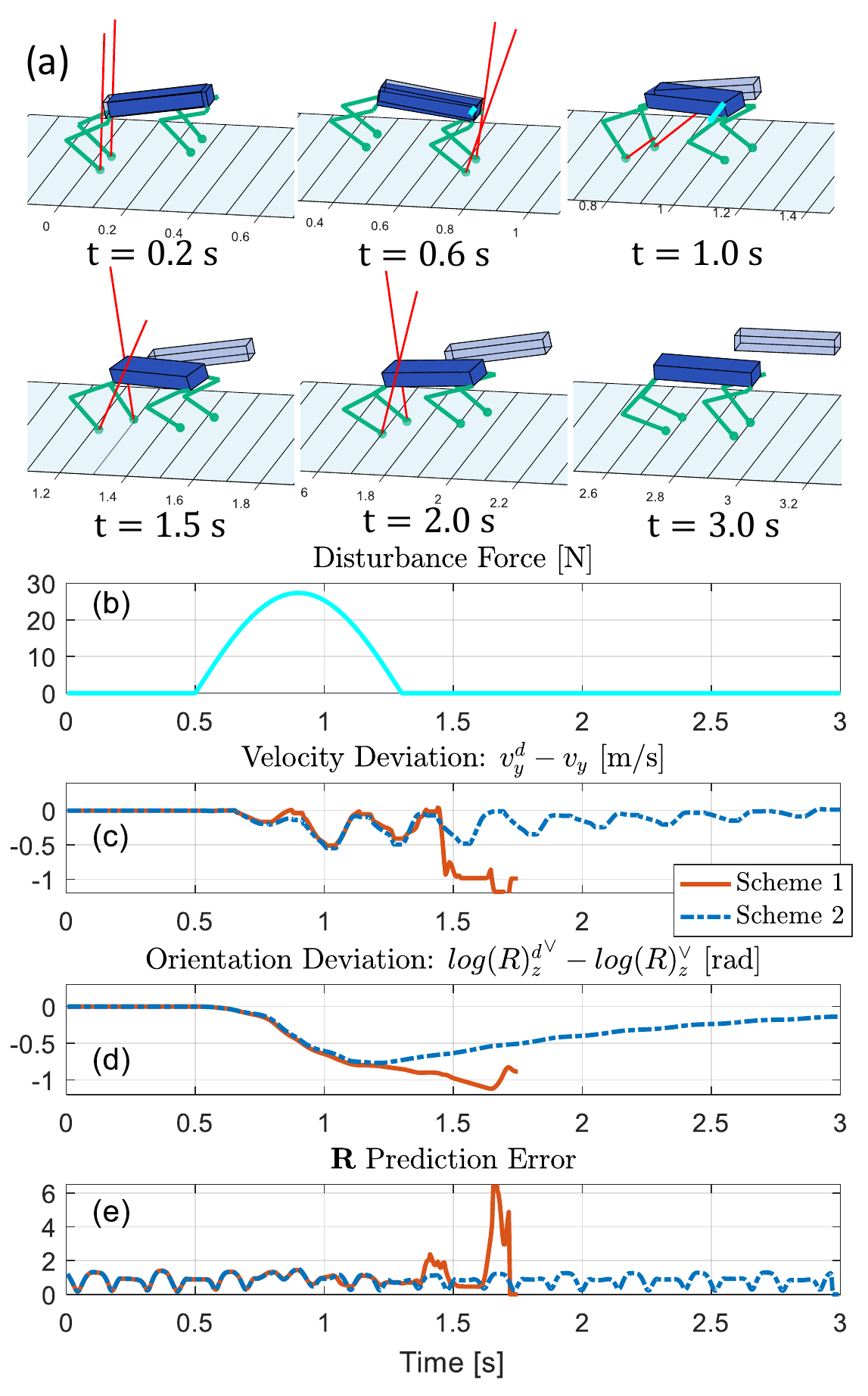}}
	\caption{Comparison between two linearization schemes. (a) A sequential snapshot of the simulation scenario for scheme 2. The GRFs are shown in red and the disturbance force is shown in cyan. The translucent box represents the reference pose of the robot. (b) Disturbance force profile. (c) Velocity deviation and (d) Orientation deviation in the $y$-direction from the reference trajectory, respectively. (e) Prediction error of the Rotation matrix.}
	\label{fig:cmp_lin}
\end{figure}

Fig. \ref{fig:cmp_lin} (a) shows a sequential snapshot of the simulated scenario for scheme 2. The GRFs are shown in red and the disturbance force (visible at t = 1.0 s and t = 1.5 s) is in cyan. Fig. \ref{fig:cmp_lin} (b) shows the disturbance force profile, which is applied at the FR shoulder of the robot. 

RF-MPC using scheme 1 fails at 1.5 s since the velocity and orientation start to diverge from the reference, as shown in Fig. \ref{fig:cmp_lin} (c) and (d), respectively. In comparison, the RF-MPC using scheme 2 recovers from the disturbance and successfully tracks the reference trajectory. To investigate the reason for the discrepancy, we defined a quantity that measures the prediction quality of the rotation matrices, 
\begin{equation}
\Phi(t) = \sum_{k=1}^N ||\widetilde{\bm{R}}_{k}-\overline{\bm{R}}_{k}||_F+||log(\bm{R}^T_{op,t+k} \widetilde{\bm{R}}_{k})^{\vee}||,
\end{equation}
where $\bm{R}_{op,t+k}\in SO(3)$ is the rotation matrix at time $t+k\cdot dt$; $\overline{\bm{R}}_k$ is the $k^{th}$ predicted rotation matrix at time $t$, whose projection to the $SO(3)$ manifold is denoted as $\widetilde{\bm{R}}_{k}\in SO(3)$; $||\cdot||_F$ is the Frobenius norm of a matrix. The prediction error of rotation matrices is plotted in Fig. \ref{fig:cmp_lin} (e), where the error value of scheme 1 became high when the system started to deviate from the reference trajectory. This numerical study serves as an empirical validation of the robustness of scheme 2 in comparison to scheme 1 in disturbance rejection.

\subsection {Reference Trajectory Generation}
\label{sec:gen_ref_traj}

\subsubsection{\textbf{Trotting}}
The reference control trajectory $\bm{u}_{k,d}$ for trotting is defined based on the heuristic that the total weight of the robot should be supported evenly by all the legs that are in contact:
\begin{equation}
\begin{aligned}
& u_{i,k,d}^{x/y} = 0\\
& u_{i,k,d}^{z} = \frac{b_{i,k}}{\sum_{i=1}^4 b_{i,k}} Mg,
\end{aligned}
\end{equation}
where $b_{i,k}\in\{0,1\}$ is a binary variable that indicates the contact condition of leg $i$ at instance $t+k$, where $b_{i,k}=1$ indicates contact phase and $0$ otherwise. The value of the binary variable $b_{i,k}$ is defined according to the time schedule of a Finite State Machine (FSM), which is introduced in Section \ref{sec:FSM}. The reference state $\bm{x}_{k,d}$ is constructed by simply assuming the robot accelerate from static pose with constant acceleration until reaching the maximal velocity. Both walking trot and running trot use the same reference trajectory.

\subsubsection{\textbf{Aperiodic Complex Dynamic Maneuver}}
The reference trajectory is generated by an off-line TO algorithm based on the single-shooting method. The twist jump motion is decomposed into three phases with fixed timing. Phase one is when four feet are in contact, which lasts for 0.1 s; phase two is when front legs lift off and hind legs are in contact, which lasts for 0.1 s; phase three is when the robot is airborne, which lasts for 0.3 s. The optimization variables are the magnitude of the GRFs, which is assumed to be constant throughout phases one and two. The cost function is the weighted sum of the control effort and the deviation from the desired landing pose. The constraints imposed in the optimizations are
\begin{itemize}
	\item Fixed initial state and bounded final state
	\item Fixed contact sequence and timing
	\item Kinematic reachability of each leg
	\item Collision avoidance with the environment
	\item Unilateral GRF stay within friction cone.
\end{itemize}
The TO is formulated as a nonlinear program with 24 variables, representing the force magnitude of 4 GRFs (each has 3 components) in two stance phases. The optimization problem is solved by the MATLAB NLP solver \textit{fmincon}.

\subsubsection{\textbf{Bounding}}
\label{sec:ref_gen_bound}
The periodic trajectory for bounding gait is generated by considering the robot as a single rigid body in 2D, which has 3 DoFs ($x,z,\theta$). The contact sequence and timing is pre-specified as front-stance, aerial phase I, hind-stance and aerial phase II. The shapes of vertical GRF and pitch torque profiles are parametrized by B\'ezier\ polynomials. Periodicity in the $z$ and $\theta$ directions is achieved by finding the scaling factors and initial condition based on the principle of impulse-scaling \cite{park2017high} analytically.

\subsubsection{\textbf{Controlled Backflip}}
To generate the reference trajectory of the controlled backflip in Section \ref{sec:backflip}, we used the open-source \textit{OptimTraj} library \cite{kelly2017introduction} to set up the TO problem using the direct collocation method. The optimization is done on a 2-D single rigid body model of the robot. The convex quadratic cost function penalizes large GRF and rewards smooth force profiles. In addition to the constraints mentioned in \textit{2)}, the following constraints are also imposed
\begin{itemize}
	\item The constraint that enforces feasible dynamics
	\item Path constraints on the state and control.
\end{itemize}
The above problem setup has 27 time steps, which results in an optimization problem with 272 variables and 366 constraints solved by MATLAB NLP solver \textit{fmincon}.

\section{Controller Implementation Details}
\label{sec:implement}
The MPC framework in Section \ref{sec:MPC_formulation} is combined with other components such as state estimation and swing leg controllers to give rise to various motions implemented on the robot hardware platform. This section presents the implementation details that are required to realize the MPC control design on the hardware.

\begin{figure*}
	\centering
	\resizebox{0.95\linewidth}{!}{\includegraphics{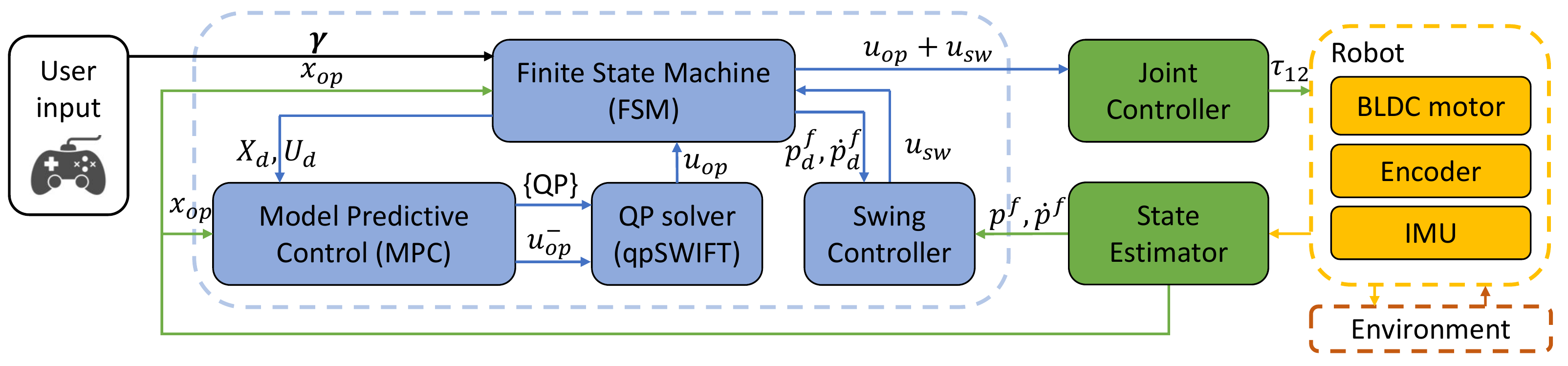}}
	\caption{Overview of the control system. The user sends commands to the on-board computer (blue), where the finite state machine schedules the gait and sends desired trajectories to the MPC block to formulate the QP. The custom QP solver qpSWIFT solves for the $\bm{u}_{op}$ and send it to the FSM. The FSM combines the stance and swing forces and send to the joint controller (green), which maps leg forces to joint torque and send to the BLDC motors. The state estimator (green) receives sensor signals for the MPC formulation of the next cycle. The previous solution of the QP $\bm{u}_{op}^{-}$ is sent to the MPC as the control at the operating point.}
	\label{fig:control_system}
\end{figure*}

Fig. \ref{fig:control_system} shows the schematics of the overall control system. The Finite State Machine (FSM) sends the desired state and control trajectories $\bm{X}_d,\bm{U}_d$ to the MPC, which formulates a Quadratic Program (QP). The QP is solved by the custom QP solver qpSWIFT\cite{pandala2019qpswift} to obtain the optimal solution $\delta{\bm{u}}$, which is added to the control at the operating point $\bm{u}_{op}^-$ to get the GRF $\bm{u}_{op}$. A swing leg controller calculates the swing force $\bm{u}_{sw}$ to track the swing foot trajectories. The commanded torque is modified by a lower-level joint controller, which compensates for friction and motor dynamics. The Brush-Less Direct Current (BLDC) motors actuate the robot to interact with the environment.

\subsection{Finite State Machine}
\label{sec:FSM}

\begin{figure}
	\centering
	\resizebox{0.85\linewidth}{!}{\includegraphics{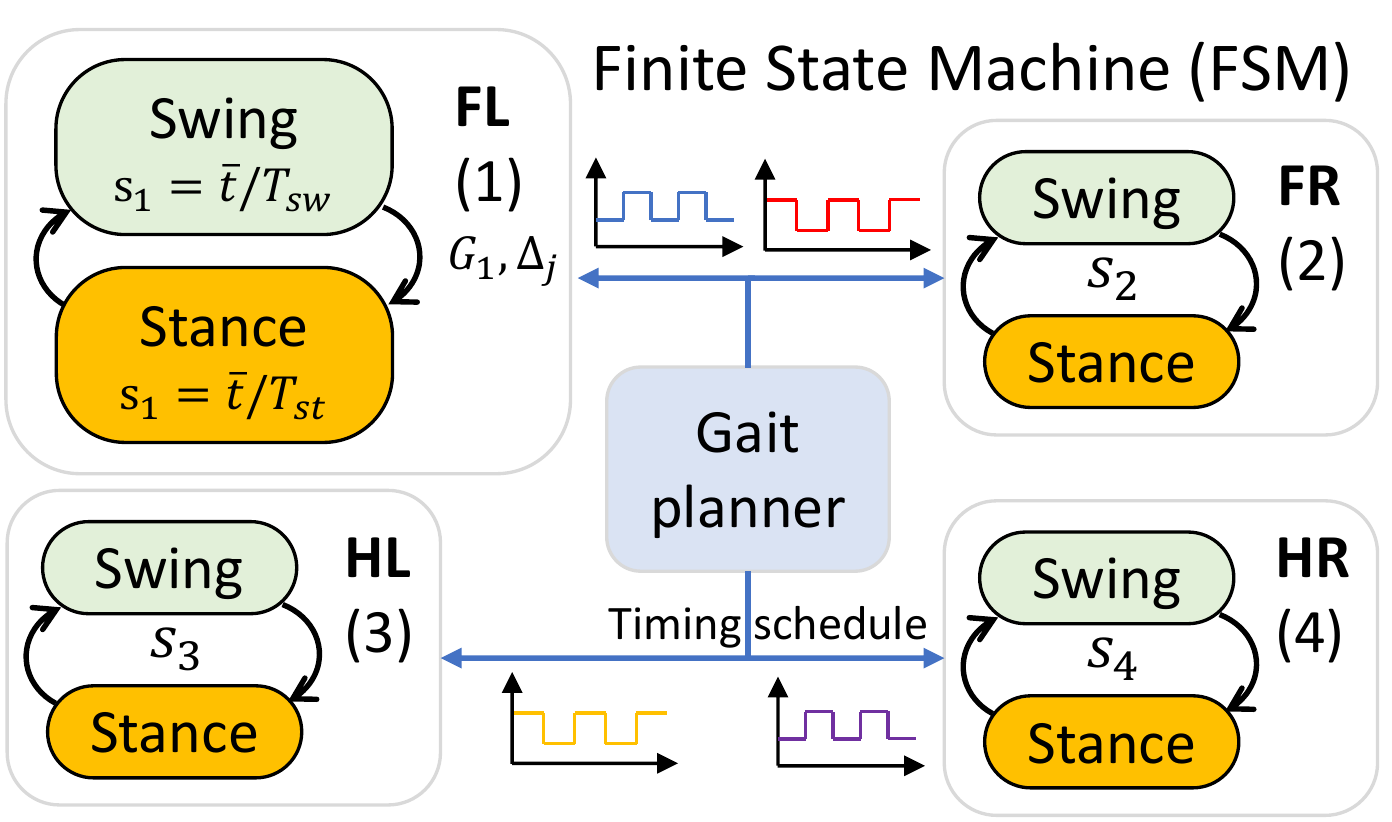}}
	\caption{Schematics of the Finite State Machine (FSM). The gait planner sends to all legs the timing schedules; the normalized variable $s_i$ is the percentile completion of the current state. $\Delta_j,j\in\{st,sw\}$ are the reset maps and $G_i$ are the guard sets.}
	\label{fig:FSM}
\end{figure}

Various gaits are generated by a finite state machine (FSM). Fig. \ref{fig:FSM} shows the schematics of the FSM where the timing schedules are sent from the gait planner to each leg. A leg independent phase variable $s_i$ quantifies the percentile completion of either stance or swing state. The phase variables are defined as $s_i:=\{\bar{t}/T_j\text{ s.t. }j\in\{st,sw\}\}$, where $\bar{t}$ represents the current dwell time, $T_{st}, T_{sw}$ are stance and swing times, respectively. The period of the gait is the sum of the dwell times $T = T_{st}+T_{sw}$. The guard sets $G_i$ and reset maps $\Delta_j$ define the transition between states. The guard sets are given as $G_i:=\{\bar{t}\text{ s.t. }\bar{t}=T_j\}$. The reset map is defined as $\Delta_j(\bar{t})=0$ such that it resets the  phase variable and current dwell time. This framework allows the implementation of any gait sequence by changing the timing schedules. The contact detection algorithm could be incorporated to adjust the gait timings and extend the time-based FSM to event-based FSM. Using the FSM scheme, trotting, bounding and aperiodic motions could be realized.

It is worth noting that the prediction horizon could cover multiple phases. Hence, in motions with aerial phases such as bounding and acrobatic jump, the RF-MPC could take into consideration of the upcoming phase change and plan the current control accordingly.

\subsection{Platform Description}
\label{sec:platform}

\begin{figure}
	\centering
	\resizebox{0.95\linewidth}{!}{\includegraphics{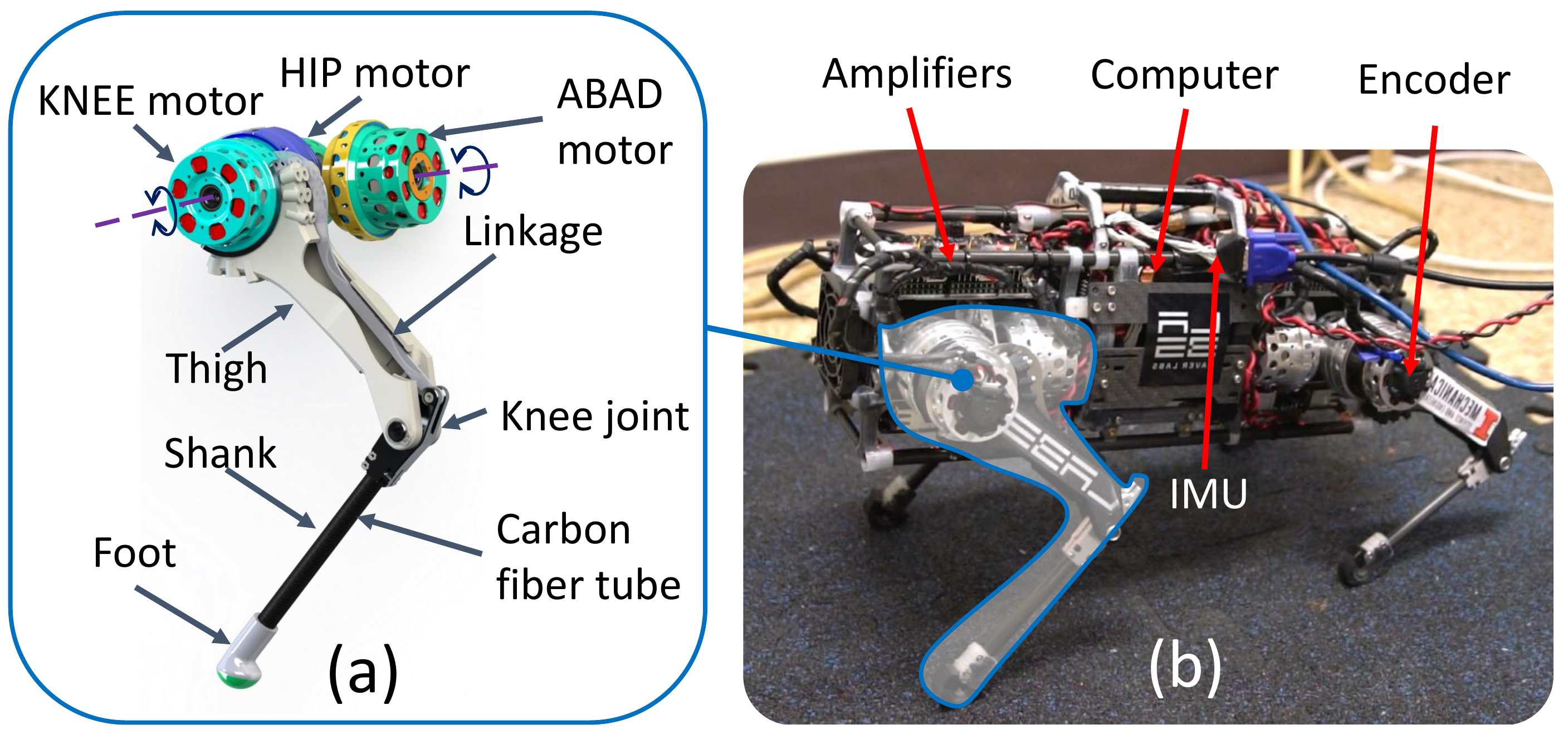}}
	\caption{An illustration of the hardware platform. (a) CAD model of the mechanical components of the leg module, which includes ABAD, HIP and KNEE modules and linkages of the leg. (b) A picture of the assembled quadruped platform, which integrates a computer, sensors including an IMU and 12 encoders.}
	\label{fig:robot_stand}
\end{figure}

The hardware platform used for the experiments is a 5.5 kg fully torque controllable, electrical quadruped robot named \textit{Panther}. Three custom-made BLDC motor units are assembled into a leg module\cite{ding2017design} which is capable of executing highly dynamic maneuvers\cite{ding2018single}\cite{ding20kino}. The body of \textit{Panther} is assembled from carbon fiber reinforced 3D printed parts that connect carbon fiber tubes and plates for higher strength to weight ratio. The point foot is cushioned with sorbothane for its shock absorption capability. The electronic system consists of an on-board computer PC104 with Intel i7-3517UE at 1.70 GHz, Elmo Gold Twitter amplifiers, RLS-RMB20 magnetic encoders on each joint, and an inertial measuring unit (IMU) VN-100. Fig. \ref{fig:robot_stand} shows the CAD model of the mechanical components of a leg module and the picture of the assembled quadruped platform with all the parts integrated.

\subsection{Computation Setup}
\label{sec:comp_setup}

The MPC framework is implemented using Simulink Real-Time (SLRT). The encoder readings and lower-level kinematics calculations are carried out at a base rate of 4 kHz, while the IMU signals are received and state estimation is performed at 1 kHz. The user input from the joystick is updated at 23 Hz, and the QP is solved at between 160 Hz to 250 Hz depending on the size of the problem. The proposed QP (\ref{eq:QP}) is solved by a custom QP solver qpSWIFT\cite{pandala2019qpswift} for all the experiments. Written in ANSI C, the solver is library-free while and it interfaces with SLRT through a gateway \textit{s-function}. A RF-MPC with prediction horizon of 6 entails solving a QP with 144 variables, 72 inequality and 72 equality constraints. 

\subsection{Swing Leg Control}
\label{sec:swing}
Since the leg mass is less than 10$\%$ of the total mass of the robot and the motor inertia is much smaller than body inertia, the inertia effect of the legs could be neglected during stance. Nevertheless, leg inertia is considered when designing the swing leg controller. The swing leg is modeled as a 3-link serial manipulator attached to a stationary base. The swing leg controller consists of feed-forward and feedback terms, where the former is based on the workspace inverse dynamics,
\begin{equation}
	\bm{\tau}_{sw}^{ff} = \bm{D}(\bm{q}) \bm{J}^{-1} (\bm{a}_x^f - \dot{\bm{J}}\dot{\bm{q}}) + \bm{h}(\bm{q},\dot{\bm{q}}),
\end{equation} 
where $\bm{\tau}_{sw}^{ff}$ is the feed-forward torque; $\bm{D}(\bm{q})$ is the Inertia matrix and $\bm{h}(\bm{q},\dot{\bm{q}})$ includes the centrifugal, Corolis and gravitational terms of the swing leg; $\bm{q},\dot{\bm{q}}$ are the joint angle and velocity vectors; $\bm{J}$ is the leg Jacobian matrix and $\dot{\bm{J}}$ is its time derivative; $\bm{a}_x^f$ is workspace acceleration vector, which is defined as
\begin{equation}
	\bm{a}_x^f = \ddot{\bm{p}}_d^f + \bm{K}_p^{ff}  (\bm{p}_d^f-\bm{p}^f) + \bm{K}_d^{ff}  (\dot{\bm{p}}_d^f - \dot{\bm{p}}^f),
\end{equation}
where $\ddot{\bm{p}}_d^f$ is the desired foot workspace acceleration; $\bm{p}_d^f, \dot{\bm{p}}_d^f$ are the desired foot position and velocity; $\bm{K}_p^{ff}$, $\bm{K}_d^{ff}$ are the position and velocity gain matrices. The full swing leg controller consists of both feed-forward and feedback terms,
\begin{equation}
	\bm{\tau}_{sw} = \bm{\tau}_{sw}^{ff} + \bm{K}_p^{fb}  (\bm{p}_d^f-\bm{p}^f) + \bm{K}_d^{fb}  (\dot{\bm{p}}_d^f - \dot{\bm{p}}^f),
\end{equation}
where $\bm{K}_p^{fb}$ and $\bm{K}_d^{fb}$ are the position and velocity gain matrices for the feedback term of the swing leg controller.

The desired foot placement is a linear combination of a velocity-based feed-forward term and a capture-point \cite{pratt2006capture} based feedback term.
\begin{equation}
	\bm{p}_{step}^f = \bm{p}^h + \frac{T_{st}}{2}  \dot{\bm{p}}_{d}^h + \sqrt{\frac{z_0^h}{g}}  (\dot{\bm{p}}^{h} - \dot{\bm{p}}_{d}^h),
\end{equation}
where $\bm{p}_{step}^f$ is the desired step location on the ground plane; $\bm{p}^h$ is the projection of the hip joint on the ground plane and $\dot{\bm{p}}^{h}$ is the corresponding velocity; $\dot{\bm{p}}_{d}^h$ is the desired hip velocity projected on the ground plane; $T_{st}$ is the stance time; $g$ is the gravitational acceleration constant; $z_0^h$ is the nominal hip height.

\subsection{State Estimation}
\label{sec:state_estimate}

Kalman Filter \cite{KalmanFilter} has been applied for a range of applications in legged robots. Meanwhile, simple linear single-input single-output (SISO) complementary filters \cite{osder1973navigation} has been proven to work robustly in practice \cite{corke2004inertial}\cite{saripalli2003tale}. The complementary filter performs low-pass filtering on a low-frequency estimation and high-pass filtering on a biased high-frequency estimation. For instance, the CoM velocity $\dot{\bm{p}}$ is obtained by combining the CoM velocity estimate from leg kinematics data $\dot{\bm{p}}^{enc}$ and the accelerometer readings $\bm{a}^{acc}$ from the on-board IMU.
\begin{equation}
	\begin{split}
		\dot{\bm{p}}_{k+1} =& ~\dot{\bm{p}}_{k} + \bm{a}_{k}\cdot \Delta t\\
		\bm{a}_{k} =& ~\bm{a}_k^{acc} - \bm{K}_p^v(\dot{\bm{p}}_k-\dot{\bm{p}}_k^{enc}),
	\end{split}
\end{equation}
where $\dot{\bm{p}}_k$ is the estimated velocity from the previous iteration; the subscript $(\cdot)_k$ is the discrete time index; $\Delta t$ is the IMU sampling period; $\bm{K}_p^v$ is a positive diagonal gain matrix; $\bm{a}_k^{acc}$ is the accelerometer reading; $\dot{\bm{p}}_k^{enc}$ is the average of all the velocities from contact feet to CoM based on kinematic calculations. Similarly, the CoM position $\bm{p}\in\mathbb{R}^3$ is estimated by fusing the CoM position estimate from leg position kinematics $\bm{p}^{enc}$ and the estimated CoM velocity $\dot{\bm{p}}$.

\subsection{Contact Detection}
\label{sec:contact_detect}

Contact sensing plays a crucial role in legged locomotion. However, conventional force estimation is fragile and noisy, which is not suitable for dynamic locomotion applications. Proprioceptive sensing \cite{wensing2017proprioceptive} is utilized in this work because of the highly-transparent actuation design. We use the generalized momenta based disturbance observer \cite{de2005sensorless}, which only requires proprioceptive measurements $\bm{q},\dot{\bm{q}}$ and the commanded torque $\bm{\tau}$. In this work, only the knee joints are considered in contact detection based on the assumption that the knee joint momentum is changed the most by the contact impact. The residual vector $\bm{r}_k$ is defined as,
\begin{equation}
	\bm{r}_k = \bm{K}_I \cdot [\bm{I}^{kn} \dot{\bm{q}}_k^{kn} - \sum_{i=1}^k(\bm{\tau}_i^{kn}+\bm{r}_{i-1})\Delta t], \bm{r}_0=0,
\end{equation}
where $\bm{r}_k\in\mathbb{R}^4$ is the residual vector for the four legs. $k$ is the index for the current instance; $\bm{K}_I$ is a diagonal gain matrix; $\bm{I}^{kn}$ is the diagonal inertia matrix for all the knee joints; $\dot{\bm{q}}_k^{kn}$ is the vector of knee joint velocity; $\bm{\tau}_k^{kn}$ is the commanded knee torque; $\bm{r}_0$ is the initial value of the residual. The summation accumulates all the previous residuals and the commanded torque. Contact is declared when the residual vector $\bm{r}_k$ exceeds a threshold value $r_{th}$.

\subsection{System Identification}
\label{sec:system_ID}

\subsubsection{\textbf{Friction Compensation}}
The gear ratio of the planetary gearbox in each motor module is 23.36:1, which is higher than other quadruped robots with proprioceptive actuation scheme, including MIT Cheetah 3 (7.67:1)\cite{bledt2018cheetah}, Mini Cheetah (6:1)\cite{katz2019mini} and Minitaur (1:1)\cite{de2018vertical}. Due to the relatively higher gear ratio, the friction induced by the gearbox and bearing is modeled and compensated for more accurate force control. Following \cite{katz2019mini}, the friction is modeled by the expression
\begin{equation}\label{eq:friction}
	\tau_{friction} = c_1\cdot\text{sat}(\omega) + c_2\cdot\tau_{motor}\cdot\text{sat}(\omega),
\end{equation}
where $\omega$ is the output angular velocity; $\tau_{motor}$ is the commanded motor torque amplified by the gear ratio; $c_1,c_2$ are tunable constants that are motor-specific. $\tau_{friction}$ is the friction compensation term and the output torque $\tau_{output}=\tau_{motor}+\tau_{friction}$. The saturation function is defined as
\begin{equation}
	\text{sat}(\omega)=
	\begin{cases}
		-1 & \omega \leq -\omega_{thr}\\
		{1}/{\omega_{thr}} &  -\omega_{thr} <\omega \leq \omega_{thr}\\
		1 & \omega_{thr} < \omega,
	\end{cases}
\end{equation}
which serves as a relaxed version of the sign function. The threshold value $\omega_{thr}$ could be tuned to prevent chattering around the equilibrium point.
\subsubsection{\textbf{Center of Mass Location}}
The CoM location estimation from the CAD model of small robots is less accurate than that of larger robots. That is because for small robots, a large portion of the body mass is occupied by the electronics, whose mass distribution cannot be exactly captured by the CAD model. Instead, we obtained the CoM location by suspending the robot by a string. When robot is stationary, the accelerometer reading is recorded. This procedure is repeated for several other known attachment points on the robot. A bundle of lines could be constructed from the readings of the accelerometer and the position of the attachment points obtained from the CAD model. The CoM location could be obtained by solving a least-square problem,
\begin{equation}
	\argmin_{\bm{p}_{CoM}}\sum_i ||\bm{p}_{CoM}-l_i||^2,
\end{equation}
where $\bm{p}_{CoM}$ is the CoM location; $l_i$ is the bundle of lines constructed from the accelerometer readings. The norm takes the smallest distance from the point to the line. The legs are commanded to a stationary nominal position throughout the experiment.

\subsubsection{\textbf{Mass Moment of Inertia}}
Mass moment of inertia $^B\bm{I}$ is an important parameter for the dynamic modeling of the robot. However, the value directly obtained from the CAD model for a small robot may not be as accurate due to the the unknown mass distribution of electronics. Therefore, a linear version of the bifilar (two-wire) torsional pendulum \cite{jardin2009optimized} is used to obtain the mass moment of inertia.

\section{Experiment Results}
\label{sec:Experiment}
The proposed RF-MPC controller is a general motion control framework which could be used to achieve multiple motion objectives. This section presents the experimental results of some common tasks for quadrupedal robots, including pose control, balancing on a moving platform, and periodic locomotive gaits such as walking trot, running trot and bounding. In addition, a controlled backflip experiment is presented to show that the RF-MPC framework is capable of controlling dynamic motions previously hard to achieve because of the presence of singularity. The gain values and the gait timing for all experiments could be found in Table \ref{tab:gains}. Clips of all the experiments could be found in the supplementary video.

\subsection{Pose and Balancing Control}
\begin{figure}
	\centering
	\resizebox{1\linewidth}{!}{\includegraphics{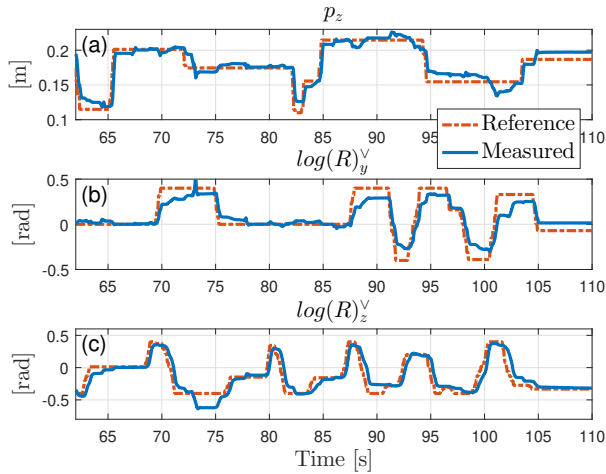}}
	\caption{Pose control experiment data. (a) the position tracking performance for $p_z$, (b) and (c) present the orientation tracking performance in the $y$ and $z$ directions.}
	\label{fig:exp_pose}
\end{figure}

\begin{figure}
	\centering
	\resizebox{0.9\linewidth}{!}{\includegraphics{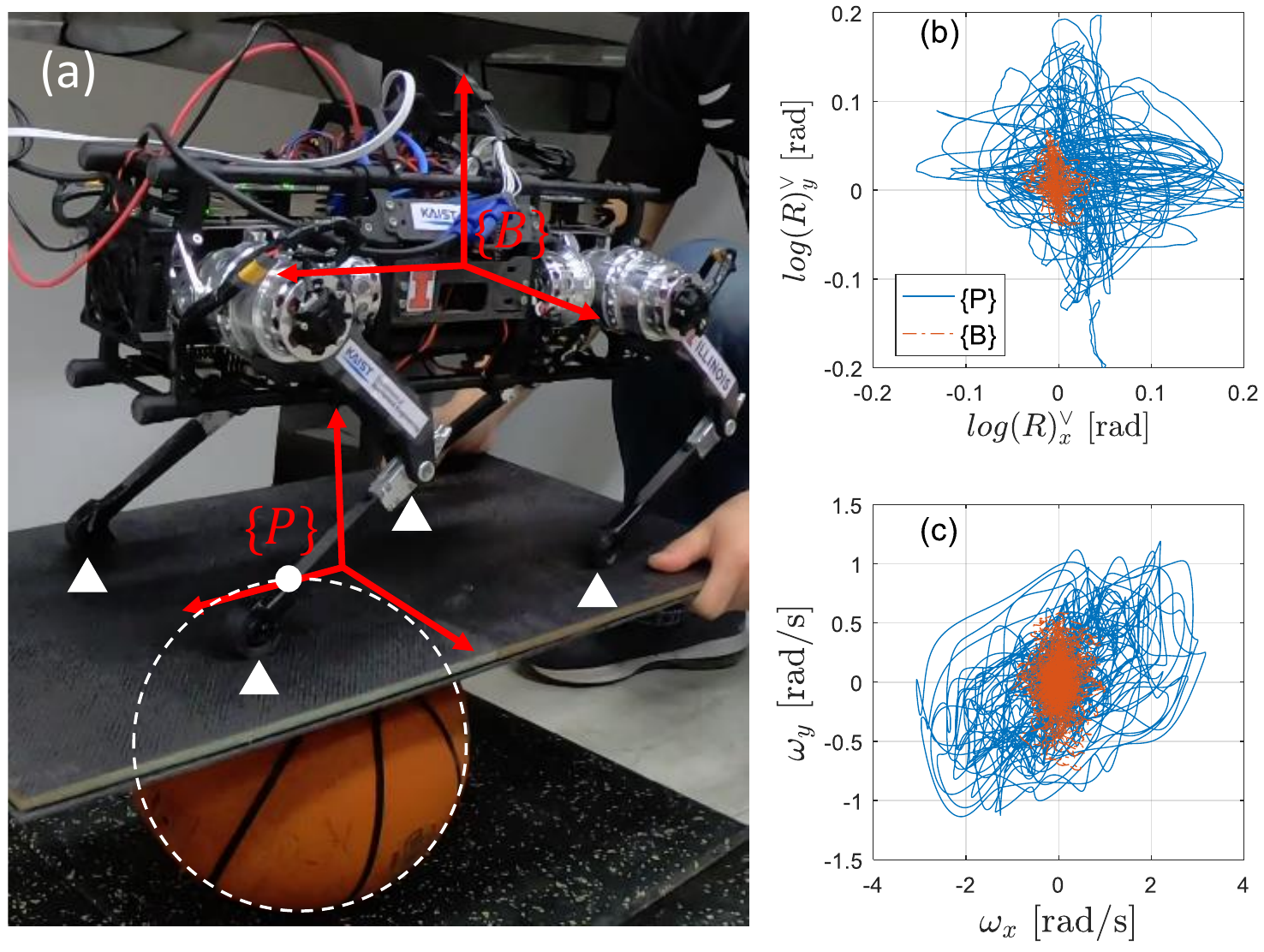}}
	\caption{The balancing control experiment (a) Experimental setup. The robot stands on a platform pivoted on a sphere, the pivot point is shown as a solid circle. The four triangles indicate the foot contact points. (b) The orientation deviation of the platform coordinate $\{P\}$ (blue) and the body coordinate $\{B\}$ (red). (c) The body angular velocity of the platform coordinate (blue) and the body coordinate (red).}
	\label{fig:exp_balance}
\end{figure}

To exhibit the tracking performance of the MPC controller, the pose control experiment is conducted. The experimenter sends the desired CoM vertical height and orientation commands in $y$ and $z$ directions to the robot from the joystick. The MPC controller continuously solves for the desired GRFs at the four feet, which are in contact with the ground throughout the experiment. The position and orientation reference tracking data shown in Fig. \ref{fig:exp_pose} suggests that RF-MPC could closely track the reference command.
To demonstrate the capability of RF-MPC to balance its body under large disturbances, the balancing experiment is presented. The experimental setup is shown in Fig. \ref{fig:exp_balance} (a). The robot stands on a pivoted platform, attempting to maintain the balance at the nominal standing pose when the platform is perturbed by the operator. The robot body coordinate $\{B\}$ and the coordinate of the platform $\{P\}$ are both plotted in Fig. \ref{fig:exp_balance} (a). The origin of $\{P\}$ is set at the center of the four feet. Fig. \ref{fig:exp_balance} (b) shows the orientation deviation in the $x$ and $y$ directions for $\{P\}$ in blue and $\{B\}$ in red; Fig. \ref{fig:exp_balance} (c) shows the angular velocity in the $x$ and $y$ directions. As shown in Fig. \ref{fig:exp_balance}, the balancing controller significantly reduces the movement of the robot's body frame $\{B\}$ compared to that of the platform-fixed frame $\{P\}$.

\subsection{Walking Trot}
\begin{figure}
	\centering
	\resizebox{1\linewidth}{!}{\includegraphics{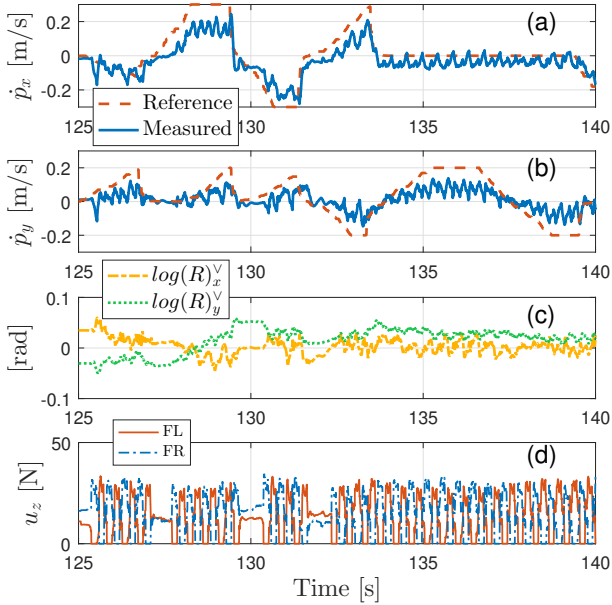}}
	\caption{Walking trot experiment data. (a) Velocity tracking in the $x$-direction.  (b) Velocity tracking in the $y$-direction (c) Orientation deviations along the $x$ and $y$-axes, where the reference is 0. (d) Vertical GRF $u_z$ for front legs.}
	\label{fig:exp_trot}
\end{figure}

To demonstrate that RF-MPC can stabilize basic locomotion gaits, the walking trot experiment is presented. The robot could move in any direction parallel to the ground while maintaining the body orientation. Fig. \ref{fig:exp_trot} (a) and (b) exhibit the velocity tracking performance of the controller, and Fig. \ref{fig:exp_trot} (c) shows that the orientation deviation is kept small (within $\pm$0.06 rad) during the walking trot experiment; Fig. \ref{fig:exp_trot} (d) presents the vertical GRF during the walking trot. The velocities are measured from the state estimation. 

\subsection{Running Trot and Bounding}
\label{sec:trotRun_pronk}

\begin{figure}
	\centering
	\resizebox{1\linewidth}{!}{\includegraphics{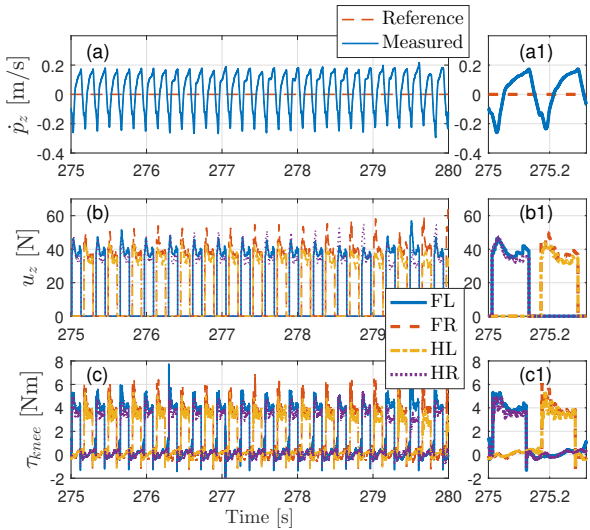}}
	\caption{Running trot experiment data. Zoomed-in views placed at the right of the figure show the details of the signals. (a) Reference and measured CoM vertical velocity in the $z$-direction (b) Vertical GRF (c) Knee torque.}
	\label{fig:exp_trotRun}
\end{figure}

To investigate the performance of RF-MPC for dynamic gaits, experiments of running trot and bounding gaits with full aerial phases are conducted. Fig. \ref{fig:exp_trotRun} presents the running trot experiment data, where Fig. \ref{fig:exp_trotRun} (a) shows that the vertical CoM velocity experiences 40 ms free fall during the aerial phase. Fig. \ref{fig:exp_trotRun} (b) shows that the robot could produce abruptly changing GRF as the contact condition changes. During the trot running experiment, the vertical GRF could reach as high as 60 N while the knee torque goes up to 6.3 Nm, as could be observed in Fig. \ref{fig:exp_trotRun}(b) and (c), respectively.

\begin{figure}
	\begin{subfigure}{0.5\textwidth}
	\centering
	\resizebox{1\linewidth}{!}{\includegraphics{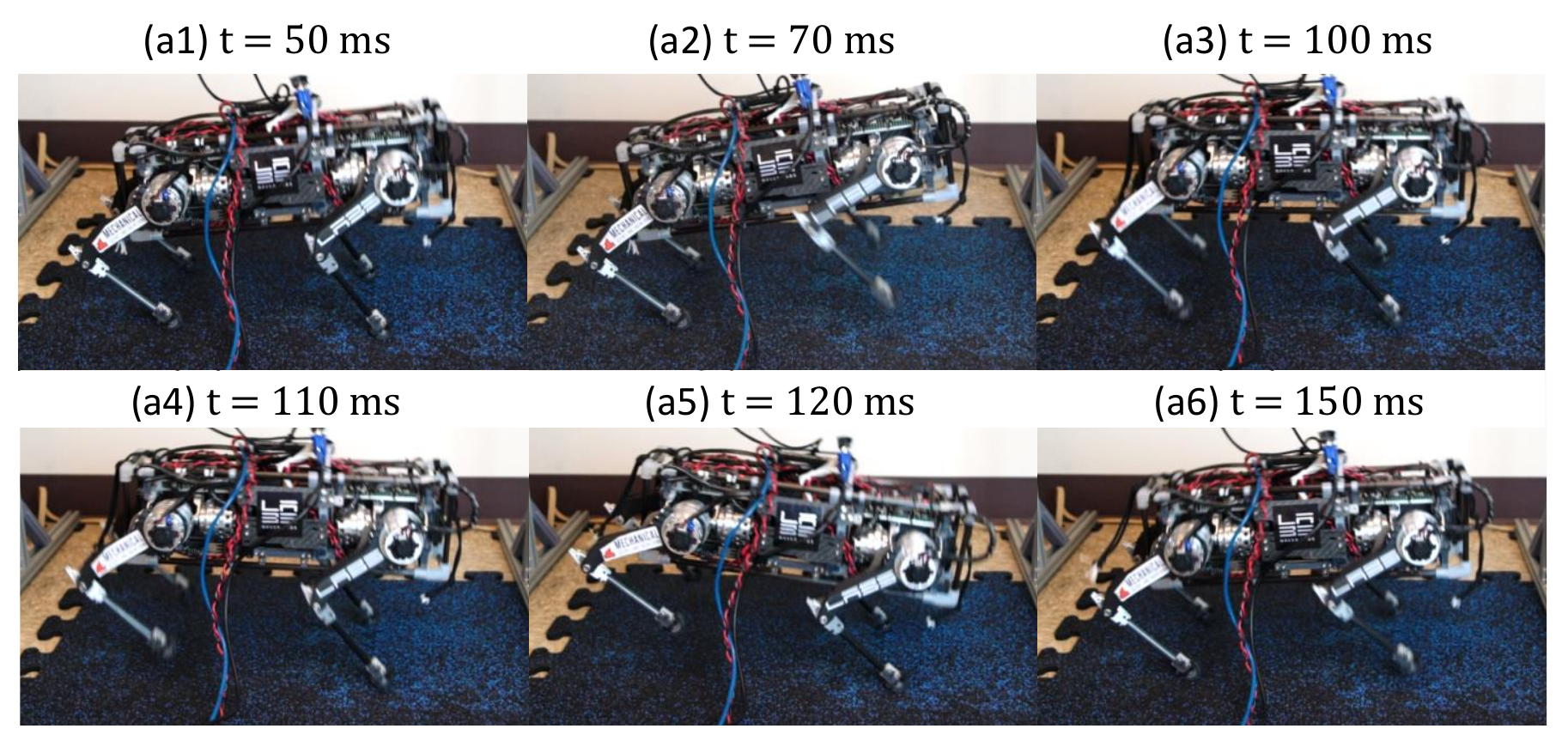}}
	\end{subfigure}
	\newline
	\begin{subfigure}{0.5\textwidth}
		\centering
		\resizebox{1\linewidth}{!}{\includegraphics{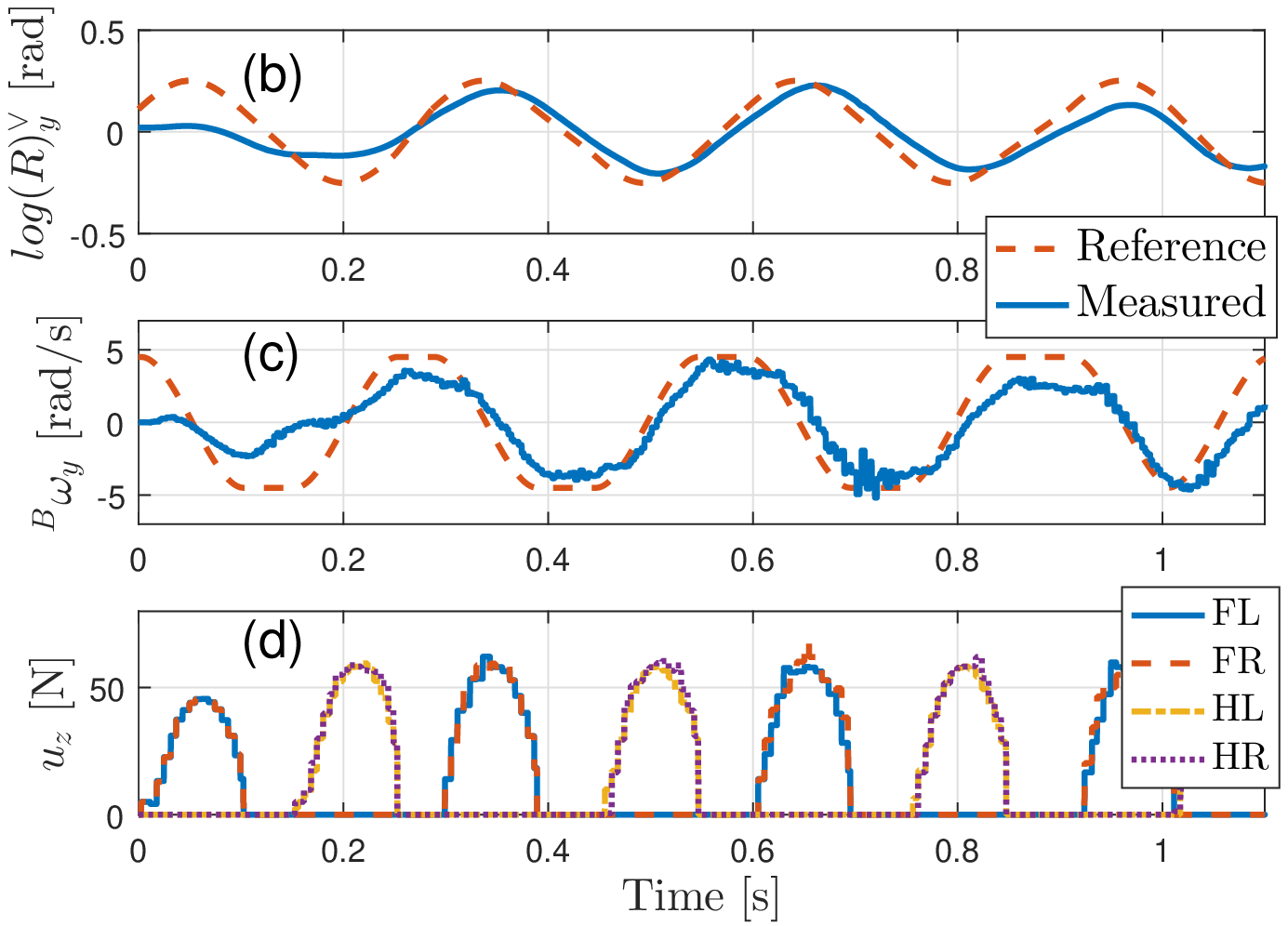}}
	\end{subfigure}
	\caption{Bounding experiment data (a) Sequential snapshots of the robot in a bounding experiment. (b) Orientation tracking in the $y$-direction $log(R)_y^{\vee}$ (c) Angular velocity tracking in the $y$-direction $^B\omega_y$ (d) Vertical GRF.}
	\label{fig:exp_bound_image_data}
\end{figure}

The bounding gait leverages the full dynamics of the robot and involves extensive body pitch oscillation. A sequential snapshots of the bounding experiment could be found in Fig. \ref{fig:exp_bound_image_data} (a). The robot starts from a static pose and the MPC stabilizes the robot to follow the desired GRF and state trajectories. More details about reference trajectory generation could be found in Section \ref{sec:gen_ref_traj}. The reference and measured trajectories of orientation and angular velocity in the $y$-direction are shown in  Fig. \ref{fig:exp_bound_image_data} (b), (c). Since the robot started from a static pose, there is an initial offset. The vertical GRF profile is shown in Fig. \ref{fig:exp_bound_image_data} (d). The transition from swing to stance phase occurs when a touchdown event is declared by the contact detection algorithm described in Section \ref{sec:contact_detect}.

\subsection{Controlled Backflip}
\label{sec:backflip}

To demonstrate the capability of RF-MPC to control dynamic maneuvers that involve singularity poses, a controlled backflip experiment is presented. As shown in Fig. \ref{fig:backflip}, the robot flips backwards around the $y$-axis, passing through the pose where the robot is upright, before it lands with the upside-down orientation. Note that even though the reference trajectory is generated based on a 2-D model of the robot as presented in Section \ref{sec:gen_ref_traj}, RF-MPC controls the 3D robot in the controlled backflip experiment without resorting to decomposition of sagittal plane motion and out-of-plane motion.

\begin{figure*}
	\centering
	\resizebox{1.0\linewidth}{!}{\includegraphics{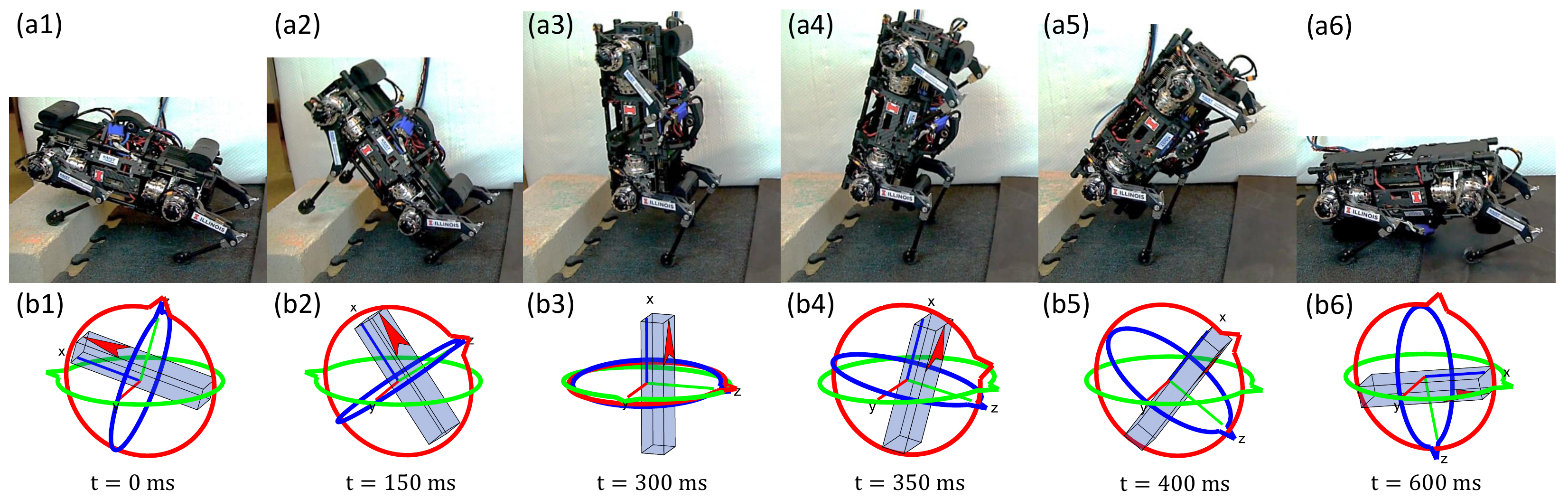}}
	\caption{ Quadruped robot \textit{Panther} performing a controlled backflip that passes through singularity pose. (a) An image sequence of the robot executing the controlled backflip, with its front legs launching from a 80 mm high platform. (b) The body orientation reconstructed from the experimental data. The rotation matrix is represented by the body coordinate frame axes ($x$-blue, $y$-red, $z$-green); the Euler angles are visualized by the three colored rings with arrow ($\phi$-blue, $\theta$-red, $\psi$-green). The robot passes through the upright pose (a3) while the hind legs are in contact with the ground. The Gimbal lock effect is shown in (b3) where axes of Euler angles are aligned.}
	\label{fig:exp_backflip_sequence}
\end{figure*}

The image sequence in Fig. \ref{fig:exp_backflip_sequence} (a) is plotted along with the body orientation, which is reconstructed from the experiment data as shown in Fig. \ref{fig:exp_backflip_sequence} (b). The rotation matrix is represented by the body coordinate frame axes ($x$-blue, $y$-red, $z$-green); the three color-coded rings correspond to the Euler angles with the roll-pitch-yaw sequence convention (roll-blue, pitch-red, yaw-green). Note that the robot is at the singular pose at around 300 ms as shown in Fig. \ref{fig:exp_backflip_sequence} (a3). In the corresponding body orientation plot, the axes of the rings for Euler angles almost coincide. Therefore, the robot indeed passes through the singularity pose when the RF-MPC is actively controlling the hind legs to track the reference trajectory. To the best of the authors' knowledge, this is the first instance of hardware experimental implementation of MPC to control acrobatic motion which involves singularity.

\begin{figure}
	\centering
	\resizebox{1.0\linewidth}{!}{\includegraphics{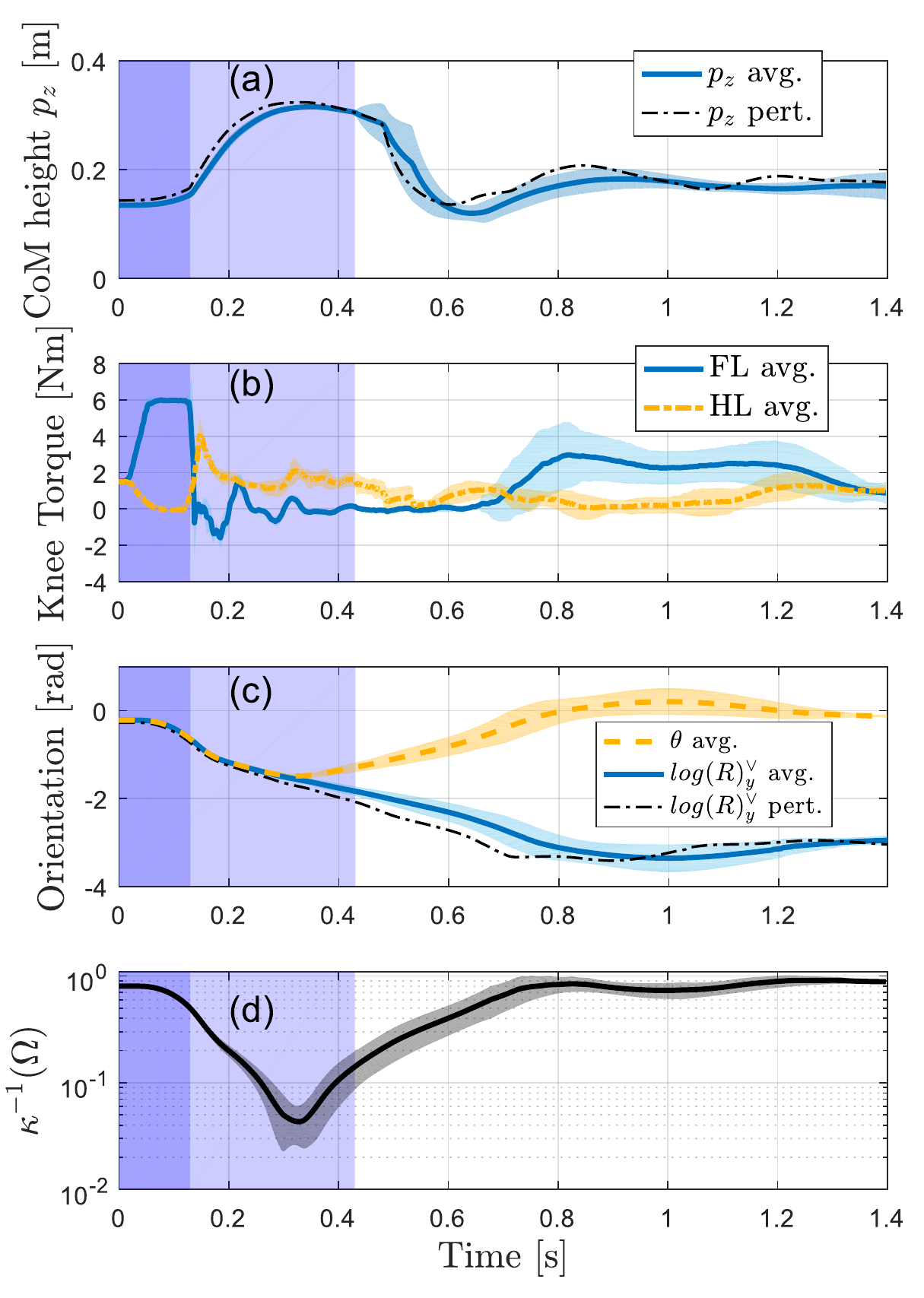}}
	\caption{Experimental data from 10 controlled backflip trials. The solid lines are the average (avg.) of all the tests, and the shaded tube is the range within one standard deviation. The black dash-dot curves in (a) and (c) are from the case where the initial condition of the backflip is perturbed (pert.). The deep-shaded area is when four legs are in contact; the light-shaded area is when hind legs are in contact, and the non-shaded area is when all legs are in the impedance control phase. (a) The CoM height. (b) The knee torque of legs FL and HL. (c) Comparison between the pitch angle $\theta$ and rotation matrix $log(R)_y^{\vee}$. (d) The function $\kappa^{-1}(\mathcal{T}_{\bm{\Theta}})$ indicates that the robot indeed encountered the singular pose in the controlled backflip experiment.}
	\label{fig:exp_backflip}
\end{figure}

Fig. \ref{fig:exp_backflip} presents the data from the controlled backflip experiment, where the robot goes through three phases. The deep-shaded area corresponds to the phase when all four legs are in contact; the light-shaded area indicates the phase when only the hind legs are in contact; the non-shaded area corresponds to the landing phase. The RF-MPC controller is activated during the first two phases, and an impedance control is utilized in the third phase. Experiment data gathered from 10 backflip trials are shown in Fig. \ref{fig:exp_backflip}, where the solid lines are the mean values of all the tests, and the shaded tube is the value within one standard deviation. 

As could be observed from Fig. \ref{fig:exp_backflip} (d), the robot passes through the neighborhood of singularity as the number introduced in Section \ref{sec:singularity_metric} drops below the threshold $10^{-1}$. Fig. \ref{fig:exp_backflip} (c) shows that the pitch angle $\theta$ is not monotonic throughout the controlled backflip while $log(R)_y^{\vee}$ decreases monotonically. The dash-dot curves in Fig. \ref{fig:exp_backflip} (a) and (c) are from the experiment trial where the initial state of the robot is perturbed. Specifically, the height of the stage on which the front legs are positioned is increased from 80 mm to 130 mm. It could be observed that while in this case the trajectory of the robot deviates more than one standard deviation from the average, RF-MPC could still stabilize the motion and land safely.

\begin{table*}[t]
	\vspace{10px}		
	\caption{Cost function weights for the simulations and experiments. The values in parenthesis represent weights on the terminal costs.}
	\centering
	\small
	\setlength\tabcolsep{6pt}
	\begin{tabular}{ c c c c c c c c c c}
		\centering
						& Sim.			& Sim.  	& Sim. 		& Acro. 		& Exp. Pose/	& Exp.  		& Exp.			& Exp.			& Exp.		\\
						& Pose 			& TrotWalk  & Bound 	& Mnvr. 		& Balance   	& TrotWalk  	& TrotRun		& Bound			& Backflip	\\
		\hline
		\hline
		$Q_{p_x}$		& 3e5 (1e5)		& 1e5 		& 8e4 		& 5e6   		& 3e5 (1e5)		& 1e5  			& 1e5			& 2e5 (1.2e5)	& 1e5		\\
		$Q_{p_y}$ 		& 5e5 (1e5)		& 2e5 		& 5e4 		& 5e6 			& 5e5 (1e5)		& 1e5 (1.5e5)	& 1e5 (1.5e5)	& 4e5 			& 1e5 (2e5)	\\
		$Q_{p_z}$		& 2e5 (1e5)		& 3e5 		& 3e6  		& 5e6   		& 2e5 (1e5)		& 1.5e5 (2.2e5) & 2e4			& 1.5e5 (2e5)	& 1.5e5 (2.2e5)\\
		\hline
		$Q_{\dot{p}_x}$	& 10 (30)		& 5e2 		& 4e3(5e2) 	& 5e3   		& 10 (30)		& 1e3 (1.5e3) 	& 1e3 (1.5e3)	& 50			& 1e3 (1.5e3)\\
		$Q_{\dot{p}_y}$	& 8 (30)		& 1e3 		& 5e2 		& 5e3   		& 8  (30)		& 1e3			& 1e3			& 200 (150)		& 1e3		\\
		$Q_{\dot{p}_z}$	& 10 (30)		& 1e3 		& 7e2(5e2) 	& 5e3   		& 10 (30)		& 150 			& 100			& 30			& 150		\\
		\hline
		$Q_{R_x}$		& 5e2			& 1e3	 	& 8e3 		& 1e6  			& 5e2 			& 2e3			& 1e3 (2e3)		& 3e3 (1e3)		& 4e3 (6e3) \\
		$Q_{R_y}$		& 2e3 (3e3)		& 1e4		& 5e5 (5e4)	& 1e6   		& 2e3 (3e3)		& 2e3			& 2e3			& 4e3 (8e3)		& 0 (10)	\\
		$Q_{R_z}$		& 1e3			& 8e2		& 8e3 		& 1e6   		& 1e3			& 8e2			& 8e2			& 1e3 (3e3)		& 8e2		\\
		\hline
		$Q_{\omega_x}$	& 2				& 40 		& 2e2 		& 5e3    		& 2  			& 60 (100)		& 60 (100)		& 3 (2)			& 60 (100)	\\
		$Q_{\omega_y}$	& 4				& 40 		& 1e2 	 	& 5e3  			& 4 			& 40 (45)		& 40 (45)		& 6 (2)			& 0 (1)		\\ 
		$Q_{\omega_z}$	& 3				& 10  		& 2e2 		& 5e3   		& 3 			& 10			& 10			& 5 (8)			& 10		\\
		\hline
		$R_{u_x}$		& 0.1 			& 0.1 		& 0.2 		& 0.1 			& 0.1 			& 0.1 			& 0.1			& 0.1			& 0.1		\\
		$R_{u_y}$		& 0.1			& 0.2 		& 0.2 		& 0.1 			& 0.1 			& 0.18  		& 0.18			& 0.18			& 0.12		\\
		$R_{u_z}$		& 0.1			& 0.1 		& 0.2 		& 0.1 			& 0.1 			& 0.08  		& 0.08			& 0.2			& 0.1		\\
		\hline
		$T_{st}$		& N/A			& 0.3		& 0.1		& 0.1 (0.2) 	& N/A			& 0.3			& 0.12 / 0.2	& 0.1			& 0.12 (0.3)*\\
		$T_{sw}$		& N/A			& 0.15		& 0.16		& N/A			& N/A			& 0.15			& 0.2 / 0.1		& 0.2			& 0.3		\\
		$N_{hor}$		& 7				& 6			& 6			& 7				& 6				& 6				& 6				& 7				& 6			\\
		$\gamma$ 		& 1.0			& 1.0		& 0.9		& 0.9			& 1.0			& 1.0			& 1.0			& 0.8			& 0.8		\\
		$T_{pred}$		& 0.05			& 0.08		& 0.01		& 0.01			& 0.02			& 0.08			& 0.05			& 0.01			& 0.02		\\
		$f_{MPC}$		& 100			& 100		& 100		& 100			& 250			& 250			& 250			& 160			& 200		\\
		\hline
		\vspace{-10px}
	\end{tabular}
	\begin{tablenotes}
		\item Note: $T_{st},T_{sw}$ and $T_{pred}$ all have the unit of [s]; $N_{hor}$ is the MPC prediction horizon; $T_{pred}$ is the prediction time step; $f_{MPC}$ is the MPC control frequency with the unit of [Hz].
		\item *0.13 s is the front stance time, and 0.3 s is the hind stance time.
	\end{tablenotes}
	\label{tab:gains}
\end{table*}

\section{Discussion}
\label{sec:discussion}
In this section we briefly discuss some of the important aspects of this work. That includes the interpretation of some of the experiment results, discussion of the findings, and limitations of this work.

In this work, the proposed RF-MPC uses the rotation matrix to represent the orientation, which is capable of stabilizing dynamic motion in 3D that involves singularity in the Euler angle formulation. Specifically, Section \ref{sec:backflip} presents the controlled backflip experiment, where RF-MPC stabilized the robot to perform an acrobatic maneuver that passes through the singularity. The simulation result in Section \ref{sec:sim_acrobatic} suggests that the function value dropping below the threshold would result in the failure of the EA-MPC. Another problem EA-MPC has is shown in Fig. \ref{fig:exp_backflip} (c), where the pitch angle $\theta$ decreased until $-\frac{\pi}{2}$ and went back to $0$ rad. Notice how the monotonicity changed as the value of $\kappa^{-1}(\mathcal{T}_{\bm{\Theta}})$ went below the threshold of $10^{-1}$ s shown in Fig. \ref{fig:exp_backflip} (d). In contrast, $log(R)_y^{\vee}$ monotonically decreased to $-\pi$. Note that though the motion of the controlled backflip remains in the sagittal plane, RF-MPC is stabilizing the 3D motion without decomposing the motion into in-plane and out-of-plane parts. This experiment is our initial demonstration of the RF-MPC framework that could potentially open up the possibilities of controlling legged robots to perform the extremely agile motions as shown in \cite{dogParkour}.

One of the findings is that, a simulation case study shown in Section \ref{sec:compare_linearizations} suggests that within the RF-MPC framework, linearizing around the operating point (scheme 2) provides more robust behavior compared with linearizing around the reference trajectory (scheme 1). The result is counter-intuitive because scheme 1 uses time-varying Jacobian matrices $\bm{A}_k,\bm{B}_k$ parametrized by the reference trajectory within the prediction horizon, while scheme 2 uses matrices parameterized only by the operating point $\bm{A}|_{op},\bm{B}|_{op}$ throughout the prediction horizon. Namely, scheme 1 utilizes more information than scheme 2. Our conjecture for the reason why scheme 2 provides more robustness is stated as follows. Since RF-MPC represents orientation using the rotation matrix, which presumes $SO(3)$ structure, linearizing around the operating point guarantees an accurate dynamics model for predicted states that are close to the operating point. In comparison, when the orientation deviation from the reference trajectory is large, the dynamics linearized about the reference no longer provide a realistic local approximation of the original nonlinear dynamics. This phenomenon is shown in Fig. \ref{fig:cmp_lin}(e), where the prediction error of the rotation matrix in scheme 1 became large, which leads to the failure of the controller. The decay rate $\gamma$ is used in both schemes to discount the effect of states that are farther in the future, where the linearized model is less accurate.

A limitation of the proposed RF-MPC formulation is that the predicted rotation matrices constructed from $\bm{\xi}$ are not elements of the $SO(3)$ manifold since the first order approximation is unable to fully capture the $SO(3)$ structure. Hence, the prediction error is more pronounced for a longer prediction horizon. Currently, prediction horizon remains an important design parameter with a trade-off between the predictive ability of MPC and the accuracy of the linearized dynamics model. Prediction horizon being too long leads to inaccurate rotational dynamics, while being too short leads to myopic behaviors. To mitigate this issue, we envision a hierarchical framework with multiple MPCs running at different rates. Specifically, an MPC with simpler model and longer prediction horizon could be running at lower update rate, while the RF-MPC with shorter prediction horizon could be running at a higher rate. 

The larger deviation towards the end of the bounding motion may be caused by the simple state estimation and contact detection algorithms presented in Sections \ref{sec:state_estimate} and \ref{sec:contact_detect}, respectively. The tracking error in Fig. \ref{fig:exp_pose},  \ref{fig:exp_trot}, and \ref{fig:exp_bound_image_data} could also be affected by these reasons since the velocities were measured from state estimation instead of external sensors.

\section{Conclusion and Future Work}
\label{sec:conclusion}

In this work we presented a representation-free model predictive control framework that directly represents orientation using the rotation matrix instead of using other orientation representations. Despite the local validity of linearized dynamics on the rotation matrix, this approach introduces the possibility to stabilize 3D complex acrobatic maneuvers that involve singularities in the Euler angles formulation. By directly working on the rotation matrix, this method avoids issues arising from the usage of other representations such as unwinding phenomenon (quaternion) or singularity (Euler angles). The application of a variation-based linearization scheme and a vectorization routine linearized the nonlinear dynamics and transformed the matrix variables into vector variables. The deliberate construction of the orientation error function enabled us to formulate the MPC into the standard QP form.

We reported both simulation and experiment results of the RF-MPC controller applied on the quadruped \textit{Panther} robot. In the simulation case study presented in Section \ref{sec:compare_linearizations} we found out that in the RF-MPC framework, linearizing around the operating point provides a more robust control strategy compared with linearizing around the reference trajectory. Experiments including pose/balance control, walking/running trot and bounding were conducted on the robot. In addition, the controlled backflip experiment demonstrated that RF-MPC controller can stabilize dynamic motions that involve the singularity. By utilizing a custom QP solver qpSWIFT, the MPC could reach control frequency as high as 250 Hz.

This novel RF-MPC framework is likely to open up possibilities for quadruped robots and legged robots in general to realize extremely dynamic 3D motions. We also envision to equip the robot with special end-effectors ($e.g.$, climbing robot with claws\cite{8629057} or magnetic grippers), enabling it to climb up vertical surfaces and walk on the ceiling. Moreover, with the emergence of powerful and light-weight computing units, the variation-based formulation could potentially be applied to stabilizing acrobatic maneuvers in UAVs.

\section*{Acknowledgment}
The authors would like to thank Prof. Patrick Wensing for the insightful discussion, Prof. Jo$\tilde{\text{a}}$o Ramos for his advice and support, and Jaejun Park for his help in hardware assembly.

\ifCLASSOPTIONcaptionsoff
  \newpage
\fi

\newpage

\bibliographystyle{IEEEtran}	
\bibliography{TRO19}

\begin{thebibliography}{10}
\providecommand{\url}[1]{#1}
\csname url@samestyle\endcsname
\providecommand{\newblock}{\relax}
\providecommand{\bibinfo}[2]{#2}
\providecommand{\BIBentrySTDinterwordspacing}{\spaceskip=0pt\relax}
\providecommand{\BIBentryALTinterwordstretchfactor}{4}
\providecommand{\BIBentryALTinterwordspacing}{\spaceskip=\fontdimen2\font plus
\BIBentryALTinterwordstretchfactor\fontdimen3\font minus
  \fontdimen4\font\relax}
\providecommand{\BIBforeignlanguage}[2]{{%
\expandafter\ifx\csname l@#1\endcsname\relax
\typeout{** WARNING: IEEEtran.bst: No hyphenation pattern has been}%
\typeout{** loaded for the language `#1'. Using the pattern for}%
\typeout{** the default language instead.}%
\else
\language=\csname l@#1\endcsname
\fi
#2}}
\providecommand{\BIBdecl}{\relax}
\BIBdecl

\bibitem{ding2019real}
Y.~Ding, A.~Pandala, and H.-W. Park, ``Real-time model predictive control for
  versatile dynamic motions in quadrupedal robots,'' in \emph{2019
  International Conference on Robotics and Automation (ICRA)}.\hskip 1em plus
  0.5em minus 0.4em\relax IEEE, 2019, pp. 8484--8490.

\bibitem{aronson1982ibex}
L.~Aronson, ``The ibex—king of the cliffs,'' \emph{Massada, Tel Aviv,(in
  Hebrew)}, 1982.

\bibitem{dogParkour}
\BIBentryALTinterwordspacing
Alex \& jumpy - the parkour dog. YouTube:. [Online]. Available:
  \url{https://www.youtube.com/watch?v=39oGCTAJ9Vw&t=117s}
\BIBentrySTDinterwordspacing

\bibitem{de2018vertical}
A.~De and D.~E. Koditschek, ``Vertical hopper compositions for preflexive and
  feedback-stabilized quadrupedal bounding, pacing, pronking, and trotting,''
  \emph{The International Journal of Robotics Research}, vol.~37, no.~7, pp.
  743--778, 2018.

\bibitem{hutter2016anymal}
M.~Hutter, C.~Gehring, D.~Jud, A.~Lauber, C.~D. Bellicoso, V.~Tsounis,
  J.~Hwangbo, K.~Bodie, P.~Fankhauser, M.~Bloesch \emph{et~al.}, ``Anymal-a
  highly mobile and dynamic quadrupedal robot,'' in \emph{2016 IEEE/RSJ
  International Conference on Intelligent Robots and Systems (IROS)}.\hskip 1em
  plus 0.5em minus 0.4em\relax IEEE, 2016, pp. 38--44.

\bibitem{semini2011design}
C.~Semini, N.~G. Tsagarakis, E.~Guglielmino, M.~Focchi, F.~Cannella, and D.~G.
  Caldwell, ``Design of {HyQ}--a hydraulically and electrically actuated
  quadruped robot,'' \emph{Proceedings of the Institution of Mechanical
  Engineers, Part I: Journal of Systems and Control Engineering}, vol. 225,
  no.~6, pp. 831--849, 2011.

\bibitem{seok2013design}
S.~Seok, A.~Wang, M.~Y. Chuah, D.~Otten, J.~Lang, and S.~Kim, ``Design
  principles for highly efficient quadrupeds and implementation on the {MIT}
  cheetah robot,'' in \emph{2013 IEEE International Conference on Robotics and
  Automation}.\hskip 1em plus 0.5em minus 0.4em\relax IEEE, 2013, pp.
  3307--3312.

\bibitem{park2017high}
H.-W. Park, P.~M. Wensing, and S.~Kim, ``High-speed bounding with the
  \mbox{MIT} cheetah 2: Control design and experiments,'' \emph{The
  International Journal of Robotics Research}, vol.~36, no.~2, pp. 167--192,
  2017.

\bibitem{bledt2018cheetah}
G.~Bledt, M.~J. Powell, B.~Katz, J.~Di~Carlo, P.~M. Wensing, and S.~Kim,
  ``{MIT} cheetah 3: Design and control of a robust, dynamic quadruped robot,''
  in \emph{2018 IEEE/RSJ International Conference on Intelligent Robots and
  Systems (IROS)}.\hskip 1em plus 0.5em minus 0.4em\relax IEEE, 2018, pp.
  2245--2252.

\bibitem{8460731}
P.~{Fankhauser}, M.~{Bjelonic}, C.~{Dario Bellicoso}, T.~{Miki}, and
  M.~{Hutter}, ``Robust rough-terrain locomotion with a quadrupedal robot,'' in
  \emph{2018 IEEE International Conference on Robotics and Automation (ICRA)},
  2018, pp. 5761--5768.

\bibitem{park2020jumping}
H.-W. Park, P.~M. Wensing, and S.~Kim, ``Jumping over obstacles with mit
  cheetah 2,'' \emph{Robotics and Autonomous Systems}, p. 103703, 2020.

\bibitem{8794449}
Q.~{Nguyen}, M.~J. {Powell}, B.~{Katz}, J.~D. {Carlo}, and S.~{Kim},
  ``Optimized jumping on the mit cheetah 3 robot,'' in \emph{2019 International
  Conference on Robotics and Automation (ICRA)}, 2019, pp. 7448--7454.

\bibitem{katz2019mini}
B.~Katz, J.~Di~Carlo, and S.~Kim, ``Mini cheetah: A platform for pushing the
  limits of dynamic quadruped control,'' in \emph{2019 International Conference
  on Robotics and Automation (ICRA)}.\hskip 1em plus 0.5em minus 0.4em\relax
  IEEE, 2019, pp. 6295--6301.

\bibitem{raibert1986legged}
M.~H. Raibert, \emph{Legged robots that balance}.\hskip 1em plus 0.5em minus
  0.4em\relax MIT press, 1986.

\bibitem{righetti2013optimal}
L.~Righetti, J.~Buchli, M.~Mistry, M.~Kalakrishnan, and S.~Schaal, ``Optimal
  distribution of contact forces with inverse-dynamics control,'' \emph{The
  International Journal of Robotics Research}, vol.~32, no.~3, pp. 280--298,
  2013.

\bibitem{hutter2013hybrid}
M.~Hutter, C.~Gehring, M.~Bloesch, C.~D. Remy, and R.~Siegwart, ``Hybrid
  operational space control for compliant legged systems,'' \emph{Robotics}, p.
  129, 2013.

\bibitem{xiong2020sequential}
X.~Xiong and A.~Ames, ``Sequential motion planning for bipedal somersault via
  flywheel slip and momentum transmission with task space control,''
  \emph{arXiv preprint arXiv:2008.02432}, 2020.

\bibitem{Hwangboeaau5872}
J.~Hwangbo, J.~Lee, A.~Dosovitskiy, D.~Bellicoso, V.~Tsounis, V.~Koltun, and
  M.~Hutter, ``Learning agile and dynamic motor skills for legged robots,''
  \emph{Science Robotics}, vol.~4, no.~26, 2019.

\bibitem{herdt2010online}
A.~Herdt, H.~Diedam, P.-B. Wieber, D.~Dimitrov, K.~Mombaur, and M.~Diehl,
  ``Online walking motion generation with automatic footstep placement,''
  \emph{Advanced Robotics}, vol.~24, no. 5-6, pp. 719--737, 2010.

\bibitem{koenemann2015whole}
J.~Koenemann, A.~Del~Prete, Y.~Tassa, E.~Todorov, O.~Stasse, M.~Bennewitz, and
  N.~Mansard, ``Whole-body model-predictive control applied to the {HRP}-2
  humanoid,'' in \emph{2015 IEEE/RSJ International Conference on Intelligent
  Robots and Systems (IROS)}.\hskip 1em plus 0.5em minus 0.4em\relax IEEE,
  2015, pp. 3346--3351.

\bibitem{neunert2018whole}
M.~Neunert, M.~St{\"a}uble, M.~Giftthaler, C.~D. Bellicoso, J.~Carius,
  C.~Gehring, M.~Hutter, and J.~Buchli, ``Whole-body nonlinear model predictive
  control through contacts for quadrupeds,'' \emph{IEEE Robotics and Automation
  Letters}, vol.~3, no.~3, pp. 1458--1465, 2018.

\bibitem{grandia2019frequency}
R.~Grandia, F.~Farshidian, A.~Dosovitskiy, R.~Ranftl, and M.~Hutter,
  ``Frequency-aware model predictive control,'' \emph{IEEE Robotics and
  Automation Letters}, vol.~4, no.~2, pp. 1517--1524, 2019.

\bibitem{feedbackMPC19}
R.~{Grandia}, F.~{Farshidian}, R.~{Ranftl}, and M.~{Hutter}, ``Feedback mpc for
  torque-controlled legged robots,'' in \emph{2019 IEEE/RSJ International
  Conference on Intelligent Robots and Systems (IROS)}, 2019, pp. 4730--4737.

\bibitem{mastalli2020motion}
C.~Mastalli, I.~Havoutis, M.~Focchi, D.~G. Caldwell, and C.~Semini, ``Motion
  planning for quadrupedal locomotion: coupled planning, terrain mapping and
  whole-body control,'' \emph{IEEE Transactions on Robotics}, 2020.

\bibitem{passiveWBC20}
S.~{Fahmi}, C.~{Mastalli}, M.~{Focchi}, and C.~{Semini}, ``Passive whole-body
  control for quadruped robots: Experimental validation over challenging
  terrain,'' \emph{IEEE Robotics and Automation Letters}, vol.~4, no.~3, pp.
  2553--2560, 2019.

\bibitem{Focchi2020}
M.~Focchi, R.~Orsolino, M.~Camurri, V.~Barasuol, C.~Mastalli, D.~G. Caldwell,
  and C.~Semini, \emph{Heuristic Planning for Rough Terrain Locomotion in
  Presence of External Disturbances and Variable Perception Quality}.\hskip 1em
  plus 0.5em minus 0.4em\relax Cham: Springer International Publishing, 2020,
  pp. 165--209.

\bibitem{shuster1993survey}
M.~D. Shuster, ``A survey of attitude representations,'' \emph{Navigation},
  vol.~8, no.~9, pp. 439--517, 1993.

\bibitem{siciliano2016springer}
B.~Siciliano and O.~Khatib, \emph{Springer handbook of robotics}.\hskip 1em
  plus 0.5em minus 0.4em\relax Springer, 2016.

\bibitem{bhat1998topological}
S.~P. Bhat and D.~S. Bernstein, ``A topological obstruction to global
  asymptotic stabilization of rotational motion and the unwinding phenomenon,''
  in \emph{American Control Conference, 1998. Proceedings of the 1998},
  vol.~5.\hskip 1em plus 0.5em minus 0.4em\relax IEEE, 1998, pp. 2785--2789.

\bibitem{yang2019bee}
X.~{Yang}, Y.~{Chen}, L.~{Chang}, A.~A. {Calderón}, and N.~O.
  {Pérez-Arancibia}, ``Bee+: A 95-mg four-winged insect-scale flying robot
  driven by twinned unimorph actuators,'' \emph{IEEE Robotics and Automation
  Letters}, vol.~4, no.~4, pp. 4270--4277, 2019.

\bibitem{5991127}
C.~G. {Mayhew}, R.~G. {Sanfelice}, and A.~R. {Teel}, ``On quaternion-based
  attitude control and the unwinding phenomenon,'' in \emph{Proceedings of the
  2011 American Control Conference}, 2011, pp. 299--304.

\bibitem{bullo2004geometric}
F.~Bullo and A.~D. Lewis, \emph{Geometric control of mechanical systems:
  modeling, analysis, and design for simple mechanical control systems}.\hskip
  1em plus 0.5em minus 0.4em\relax Springer Science \& Business Media, 2004,
  vol.~49.

\bibitem{wu2015variation}
G.~Wu and K.~Sreenath, ``Variation-based linearization of nonlinear systems
  evolving on \mbox{SO(3)}and \mbox{S2},,'' \emph{IEEE Access}, vol.~3, pp.
  1592--1604, 2015.

\bibitem{lee2011stable}
T.~Lee, M.~Leok, and N.~H. McClamroch, ``Stable manifolds of saddle equilibria
  for pendulum dynamics on \mbox{S2} and \mbox{SO(3)},'' in \emph{Decision and
  Control and European Control Conference (CDC-ECC), 2011 50th IEEE Conference
  on Decision and Control and European Control Conference (CDC-ECC)}.\hskip 1em
  plus 0.5em minus 0.4em\relax IEEE, 2011, pp. 3915--3921.

\bibitem{meadows1995receding}
E.~S. Meadows, M.~A. Henson, J.~W. Eaton, and J.~B. Rawlings, ``Receding
  horizon control and discontinuous state feedback stabilization,''
  \emph{International Journal of Control}, vol.~62, no.~5, pp. 1217--1229,
  1995.

\bibitem{bock1984multiple}
H.~G. Bock and K.-J. Plitt, ``A multiple shooting algorithm for direct solution
  of optimal control problems,'' \emph{IFAC Proceedings Volumes}, vol.~17,
  no.~2, pp. 1603--1608, 1984.

\bibitem{vonStryk1993}
O.~von Stryk, \emph{Numerical Solution of Optimal Control Problems by Direct
  Collocation}.\hskip 1em plus 0.5em minus 0.4em\relax Basel: Birkh{\"a}user
  Basel, 1993, pp. 129--143.

\bibitem{full1999templates}
R.~J. Full and D.~E. Koditschek, ``Templates and anchors: neuromechanical
  hypotheses of legged locomotion on land,'' \emph{Journal of experimental
  biology}, vol. 202, no.~23, pp. 3325--3332, 1999.

\bibitem{kajita1991study}
S.~Kajita, ``Study of dynamic biped locomotion on rugged terrain-derivation and
  application of the linear inverted pendulum mode,'' in \emph{Proc. IEEE Int.
  Conf. on Robotics and Automation, Sacramento, CA, 1991}, 1991, pp.
  1405--1411.

\bibitem{ramos2018humanoid}
J.~Ramos and S.~Kim, ``Humanoid dynamic synchronization through whole-body
  bilateral feedback teleoperation,'' \emph{IEEE Transactions on Robotics},
  vol.~34, no.~4, pp. 953--965, 2018.

\bibitem{xiong2019orbit}
X.~{Xiong} and A.~D. {Ames}, ``Orbit characterization, stabilization and
  composition on 3d underactuated bipedal walking via hybrid passive linear
  inverted pendulum model,'' in \emph{2019 IEEE/RSJ International Conference on
  Intelligent Robots and Systems (IROS)}, 2019, pp. 4644--4651.

\bibitem{Orin2013a}
D.~E. Orin, A.~Goswami, and S.~H. Lee, ``{Centroidal dynamics of a humanoid
  robot},'' \emph{Autonomous Robots}, vol.~35, no. 2-3, pp. 161--176, 2013.

\bibitem{dai2014whole}
H.~Dai, A.~Valenzuela, and R.~Tedrake, ``Whole-body motion planning with simple
  dynamics and full kinematics,'' in \emph{Proceedings of the IEEE-RAS
  international conference on humanoid robots}, 2014.

\bibitem{wensing2013generation}
P.~M. Wensing and D.~E. Orin, ``Generation of dynamic humanoid behaviors
  through task-space control with conic optimization,'' in \emph{2013 IEEE
  International Conference on Robotics and Automation}.\hskip 1em plus 0.5em
  minus 0.4em\relax IEEE, 2013, pp. 3103--3109.

\bibitem{li2020centroidal}
C.~Li, Y.~Ding, and H.-W. Park, ``Centroidal-momentum-based trajectory
  generation for legged locomotion,'' \emph{Mechatronics}, vol.~68, p. 102364,
  2020.

\bibitem{marsden1995introduction}
J.~E. Marsden and T.~S. Ratiu, ``Introduction to mechanics and symmetry,''
  \emph{Physics Today}, vol.~48, no.~12, p.~65, 1995.

\bibitem{VBL_QP}
M.~{Chignoli} and P.~M. {Wensing}, ``Variational-based optimal control of
  underactuated balancing for dynamic quadrupeds,'' \emph{IEEE Access}, vol.~8,
  pp. 49\,785--49\,797, 2020.

\bibitem{lee2010geometric}
T.~Lee, M.~Leoky, and N.~H. McClamroch, ``Geometric tracking control of a
  quadrotor uav on \mbox{SE(3)},'' in \emph{Decision and Control (CDC), 2010
  49th IEEE Conference on}.\hskip 1em plus 0.5em minus 0.4em\relax IEEE, 2010,
  pp. 5420--5425.

\bibitem{featherstone2014rigid}
R.~Featherstone, \emph{Rigid body dynamics algorithms}.\hskip 1em plus 0.5em
  minus 0.4em\relax Springer, 2014.

\bibitem{graham2018kronecker}
A.~Graham, \emph{Kronecker products and matrix calculus with
  applications}.\hskip 1em plus 0.5em minus 0.4em\relax Courier Dover
  Publications, 2018.

\bibitem{trinkle1997dynamic}
J.~C. Trinkle, J.-S. Pang, S.~Sudarsky, and G.~Lo, ``On dynamic
  multi-rigid-body contact problems with coulomb friction,'' \emph{ZAMM-Journal
  of Applied Mathematics and Mechanics/Zeitschrift f{\"u}r Angewandte
  Mathematik und Mechanik}, vol.~77, no.~4, pp. 267--279, 1997.

\bibitem{5153127}
Y.~{Wang} and S.~{Boyd}, ``Fast model predictive control using online
  optimization,'' \emph{IEEE Transactions on Control Systems Technology},
  vol.~18, no.~2, pp. 267--278, March 2010.

\bibitem{boyd2004convex}
S.~Boyd and L.~Vandenberghe, \emph{Convex optimization}.\hskip 1em plus 0.5em
  minus 0.4em\relax Cambridge university press, 2004.

\bibitem{craig2009introduction}
J.~J. Craig, \emph{Introduction to robotics: mechanics and control, 3/E}.\hskip
  1em plus 0.5em minus 0.4em\relax Pearson Education India, 2009.

\bibitem{di2018dynamic}
J.~Di~Carlo, P.~M. Wensing, B.~Katz, G.~Bledt, and S.~Kim, ``Dynamic locomotion
  in the {MIT} {Cheetah} 3 through convex model-predictive control,'' in
  \emph{2018 IEEE/RSJ International Conference on Intelligent Robots and
  Systems (IROS)}.\hskip 1em plus 0.5em minus 0.4em\relax IEEE, 2018, pp. 1--9.

\bibitem{miniGithub}
MIT-Biomimetics-Robotics-Lab, ``Cheetah-software,''
  \url{https://github.com/charlespwd/project-title}[Accessed 29 June 2020],
  2019.

\bibitem{kelly2017introduction}
M.~Kelly, ``An introduction to trajectory optimization: How to do your own
  direct collocation,'' \emph{SIAM Review}, vol.~59, no.~4, pp. 849--904, 2017.

\bibitem{pandala2019qpswift}
A.~G. Pandala, Y.~Ding, and H.-W. Park, ``qp{SWIFT}: A real-time sparse
  quadratic program solver for robotic applications,'' \emph{IEEE Robotics and
  Automation Letters}, vol.~4, no.~4, pp. 3355--3362, 2019.

\bibitem{ding2017design}
Y.~Ding and H.-W. Park, ``Design and experimental implementation of a
  quasi-direct-drive leg for optimized jumping,'' in \emph{Intelligent Robots
  and Systems (IROS), 2017 IEEE/RSJ International Conference}.\hskip 1em plus
  0.5em minus 0.4em\relax IEEE, 2017, pp. 300--305.

\bibitem{ding2018single}
Y.~Ding, C.~Li, and H.-W. Park, ``Single leg dynamic motion planning with
  mixed-integer convex optimization,'' in \emph{2018 IEEE/RSJ International
  Conference on Intelligent Robots and Systems (IROS)}.\hskip 1em plus 0.5em
  minus 0.4em\relax IEEE, 2018, pp. 1--6.

\bibitem{ding20kino}
------, ``Kinodynamic motion planning for multi-legged robot jumping via
  mixed-integer convex program,'' in \emph{2020 IEEE/RSJ International
  Conference on Intelligent Robots and Systems (IROS)}.

\bibitem{pratt2006capture}
J.~Pratt, J.~Carff, S.~Drakunov, and A.~Goswami, ``Capture point: A step toward
  humanoid push recovery,'' in \emph{Humanoid Robots, 2006 6th IEEE-RAS
  International Conference on}.\hskip 1em plus 0.5em minus 0.4em\relax IEEE,
  2006, pp. 200--207.

\bibitem{KalmanFilter}
R.~E. Kalman, ``A new approach to linear filtering and prediction problems,''
  \emph{Transactions of the ASME--Journal of Basic Engineering}, vol.~82, no.
  Series D, pp. 35--45, 1960.

\bibitem{osder1973navigation}
S.~Osder, W.~Rouse, and L.~Young, ``Navigation, guidance, and control systems
  for {V/STOL} aircraft.'' \emph{Sperry Tech}, vol.~1, no.~3, 1973.

\bibitem{corke2004inertial}
P.~Corke, ``An inertial and visual sensing system for a small autonomous
  helicopter,'' \emph{Journal of robotic systems}, vol.~21, no.~2, pp. 43--51,
  2004.

\bibitem{saripalli2003tale}
S.~Saripalli, J.~M. Roberts, P.~Corke, G.~Buskey, and G.~Sukhatme, ``A tale of
  two helicopters,'' in \emph{Proceedings 2003 IEEE/RSJ International
  Conference on Intelligent Robots and Systems, 2003 (IROS 2003)},
  vol.~1.\hskip 1em plus 0.5em minus 0.4em\relax IEEE, 2003, pp. 805--810.

\bibitem{wensing2017proprioceptive}
P.~M. Wensing, A.~Wang, S.~Seok, D.~Otten, J.~Lang, and S.~Kim,
  ``Proprioceptive actuator design in the {MIT} {Cheetah}: Impact mitigation
  and high-bandwidth physical interaction for dynamic legged robots,''
  \emph{IEEE Transactions on Robotics}, vol.~33, no.~3, pp. 509--522, 2017.

\bibitem{de2005sensorless}
A.~De~Luca and R.~Mattone, ``Sensorless robot collision detection and hybrid
  force/motion control,'' in \emph{Proceedings of the 2005 IEEE international
  conference on robotics and automation}.\hskip 1em plus 0.5em minus
  0.4em\relax IEEE, 2005, pp. 999--1004.

\bibitem{jardin2009optimized}
M.~R. Jardin and E.~R. Mueller, ``Optimized measurements of
  unmanned-air-vehicle mass moment of inertia with a bifilar pendulum,''
  \emph{Journal of Aircraft}, vol.~46, no.~3, pp. 763--775, 2009.

\bibitem{8629057}
J.~{Park}, D.~H. {Kong}, and H.~{Park}, ``Design of anti-skid foot with passive
  slip detection mechanism for conditional utilization of heterogeneous foot
  pads,'' \emph{IEEE Robotics and Automation Letters}, vol.~4, no.~2, pp.
  1170--1177, 2019.

\end{thebibliography}

\begin{IEEEbiography}[{\includegraphics[width=1in,height=1.25in,clip,keepaspectratio]{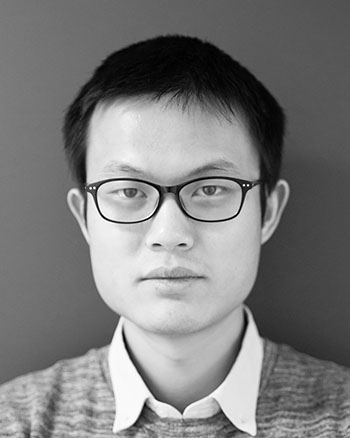}}]{Yanran Ding}
(M'16) received his B.S. degree in Mechanical Engineering from Shanghai Jiao Tong University (SJTU), Shanghai, China, in 2015 and the M.S. degree from the Mechanical Science and Engineering Department, University of Illinois at Urbana-Champaign (UIUC), Champaign, in 2017. He is currently pursuing his Ph.D. degree at the Dynamic Robotics Laboratory in UIUC. His research interests include the design of agile robotic systems and optimization-based control for legged robots to achieve dynamic motions. He is one of the best student paper finalists in the International Conference on Intelligent Robots and Systems (IROS) 2017.
\end{IEEEbiography}
\begin{IEEEbiography}[{\includegraphics[width=1in,height=1.25in,clip,keepaspectratio]{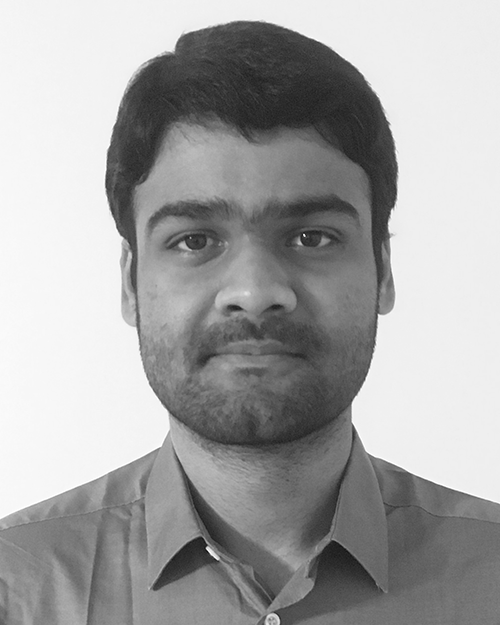}}]{Abhishek Pandala}
 received his Master of Science in Mechanical Engineering from the University of Illinois at Urbana-Champaign (UIUC) in 2019 and a Dual Degree (B.Tech and M.Tech) in Mechanical Engineering from the Indian Institute of Technology Madras (IIT-M) in 2017. He is currently pursuing his Ph.D. degree in Mechanical Engineering at the Virginia Polytechnic Institute and State University. His research interests include optimization-based control of dynamical systems with application to high degree of freedom robots.
\end{IEEEbiography}
\begin{IEEEbiography}[{\includegraphics[width=1in,height=1.25in,clip,keepaspectratio]{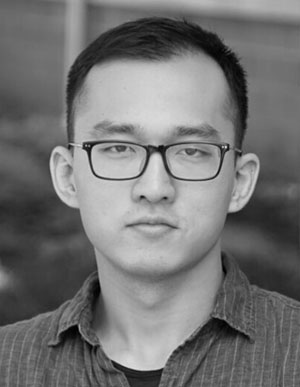}}]{Chuanzheng Li}
 received his B.S. degree in Mechatronics from Zhejiang University, Hangzhou, China in 2014, and the M.S. degree from the Mechanical Science and Engineering Department, University of Illinois at Urbana-Champaign, Champaign, IL, USA in 2017. He is currently in the Ph.D. program at University of Illinois at Urbana-Champaign supervised by Dr. Hae-Won Park, working primarily on the design of mechatronic systems and the real-time control of legged robots.
\end{IEEEbiography}
\begin{IEEEbiography}[{\includegraphics[width=1in,height=1.25in,clip,keepaspectratio]{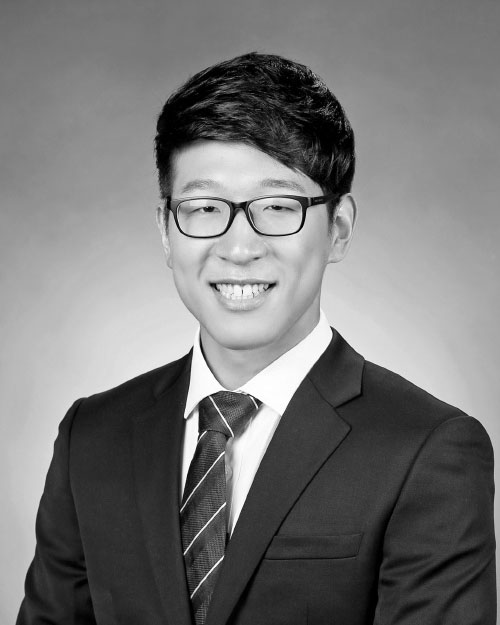}}]{Young-Ha Shin}
 received his BS degrees in the Department of Mechanical Engineering from Korea Advanced Institute of Science and Technology (KAIST), Daejeon, South Korea in 2019. He is currently a graduate student in the MS course in the Department of Mechanical Engineering in Korea Advanced Institute of Science and Technology (KAIST), Daejeon, South Korea. His research interests include actuator design, model predictive control for quadruped robots.
\end{IEEEbiography}
\begin{IEEEbiography}[{\includegraphics[width=1in,height=1.25in,clip,keepaspectratio]{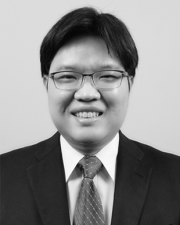}}]{Hae-Won Park}
 is an Assistant Professor of Mechanical Engineering at the Korea Advanced Institute of Science and Technology (KAIST). He received B.S. and M.S. degrees from Yonsei University, Seoul, Korea, in 2005 and 2007, respectively, and the Ph.D. degree from the University of Michigan, in 2012, all in mechanical engineering. His research interests lie at the intersection of control, dynamics, and mechanical design of robotic systems, with special emphasis on legged locomotion robots. He is the recipient of the 2018 National Science Foundation (NSF) CAREER Award, NSF most prestigious awards in support of early-career faculty.
\end{IEEEbiography}

\end{document}